\documentclass[letterpaper]{article} % DO NOT CHANGE THIS
\usepackage[]{aaai2026}  % DO NOT CHANGE THIS
\usepackage{times}  % DO NOT CHANGE THIS
\usepackage{helvet}  % DO NOT CHANGE THIS
\usepackage{courier}  % DO NOT CHANGE THIS
\usepackage[hyphens]{url}  % DO NOT CHANGE THIS
\usepackage{graphicx} % DO NOT CHANGE THIS
\usepackage{natbib}  % DO NOT CHANGE THIS AND DO NOT ADD ANY OPTIONS TO IT
\usepackage{caption} % DO NOT CHANGE THIS AND DO NOT ADD ANY OPTIONS TO IT
\usepackage{algorithm}
\usepackage{algorithmic}

\usepackage{newfloat}
\usepackage{listings}

\usepackage{amsmath,amssymb,amsfonts} % For math
\usepackage{booktabs}      % For professional-quality tables
\usepackage{nicefrac}      % For compact symbols like 1/2, etc.
\usepackage{microtype}      % For microtypography
\usepackage{subcaption}     % For subfigures
\usepackage{multirow}       % Required for multirow cells in tables
\usepackage{siunitx}        % For aligning numbers by decimal point
\usepackage{array}          % For custom column types in tables
\usepackage{makecell}

\usepackage{tcolorbox}
\tcbuselibrary{most}

\newcommand{\ie}{\textit{i.e.}}
\newcommand{\eg}{\textit{e.g.}}

\DeclareCaptionStyle{ruled}{labelfont=normalfont,labelsep=colon,strut=off} % DO NOT CHANGE THIS
\floatstyle{ruled}
\newfloat{listing}{tb}{lst}{}
\floatname{listing}{Listing}
\title{Discovering Decoupled Function Modules in Large Language Models}
\author{
    Yanke Yu\textsuperscript{\rm 1}, Jin Li\textsuperscript{\rm 1}, Ying Sun\textsuperscript{\rm 1}\thanks{Corresponding author.}, Ping Li\textsuperscript{\rm 2}, Zhefeng Wang\textsuperscript{\rm 2}, Yi Zheng\textsuperscript{\rm 2}
}
\affiliations{
    \textsuperscript{\rm 1}Thrust of AI, Hong Kong University of Science and Technology (Guangzhou)\\
    \textsuperscript{\rm 2}Huawei Technologies Ltd.\\
    rankyurky@gmail.com,
    jli740@connect.hkust-gz.edu.cn,
    yings@hkust-gz.edu.cn \\
    \{liping61, wangzhefeng, zhengyi29\}@huawei.com
    
}

\usepackage{bibentry}
\begin{document}

\maketitle

\begin{abstract}

Understanding the internal functional organization of Large Language Models (LLMs) is crucial for improving their trustworthiness and performance.
However, how LLMs organize different functions into modules remains highly unexplored.
To bridge this gap, we formulate a function module discovery problem and propose an Unsupervised LLM Cross-layer MOdule Discovery (ULCMOD) framework that simultaneously disentangles the large set of neurons in the entire LLM into modules while discovering the topics of input samples related to these modules.
Our framework introduces a novel objective function and an efficient Iterative Decoupling (IterD) algorithm.
Extensive experiments show that our method discovers high-quality, disentangled modules that capture more meaningful semantic information and achieve superior performance in various downstream tasks.
Moreover, our qualitative analysis reveals that the discovered modules show function comprehensiveness, function hierarchy, and clear function spatial arrangement within LLMs.
Our work provides a novel tool for interpreting LLMs' function modules, filling a critical gap in LLMs' interpretability research.

\end{abstract}

\begin{links}
    \vspace{-0.2cm}
    \link{Code/Appendix}{https://github.com/rank-Yu/llm-modules}
    \vspace{-0.4cm}
\end{links}

\section{Introduction}
\label{introduction}

Large Language Models (LLMs) have demonstrated remarkable human-like capabilities, such as mathematical problem-solving, coding, and information seeking~\cite{zhao2023survey, gong2024graph, xin2025llmcdsr}. However, the internal functional mechanisms of how LLMs perform complicated tasks remain unclear. A deeper understanding is crucial to improve trustworthiness, support diagnosis of model performance, and benefit potential improvements in model capabilities~\cite{huang2025survey, xu2024hallucination, yao2023editing, wang2024knowledge, guo2025towards}.

Recent studies have examined LLM internal patterns from various perspectives, such as neuron activation patterns~\cite{bills2023language, voita2023neurons}, how models represent abstract concepts~\cite{bricken2023monosemanticity, templeton2024scaling, lindsey2025biology}, or to analyze computational circuits~\cite{hanna2023does, conmy2304towards, ameisen2025circuit}. Interestingly, some studies have shown evidence for diversified activation patterns for different functions in LLMs. For example, \citet{wang2022finding, panigrahi2023task, zhao2023unveiling, tang2024language, zhao2024large} analyzed activated neurons for inputs with different predefined topic labels and discovered distinct distributions among these neurons. \citet{zhang2023emergent, xiao2024configurable} found that specific neurons are responsible for distinct functions. \citet{templeton2024scaling} used Sparse AutoEncoder to learn dictionary features for LLM activations, finding that the embeddings of the dictionary features form clusters that align with their semantics.

\begin{figure}
    \vspace{-0.2cm}
    \centering
    \includegraphics[width=\linewidth]{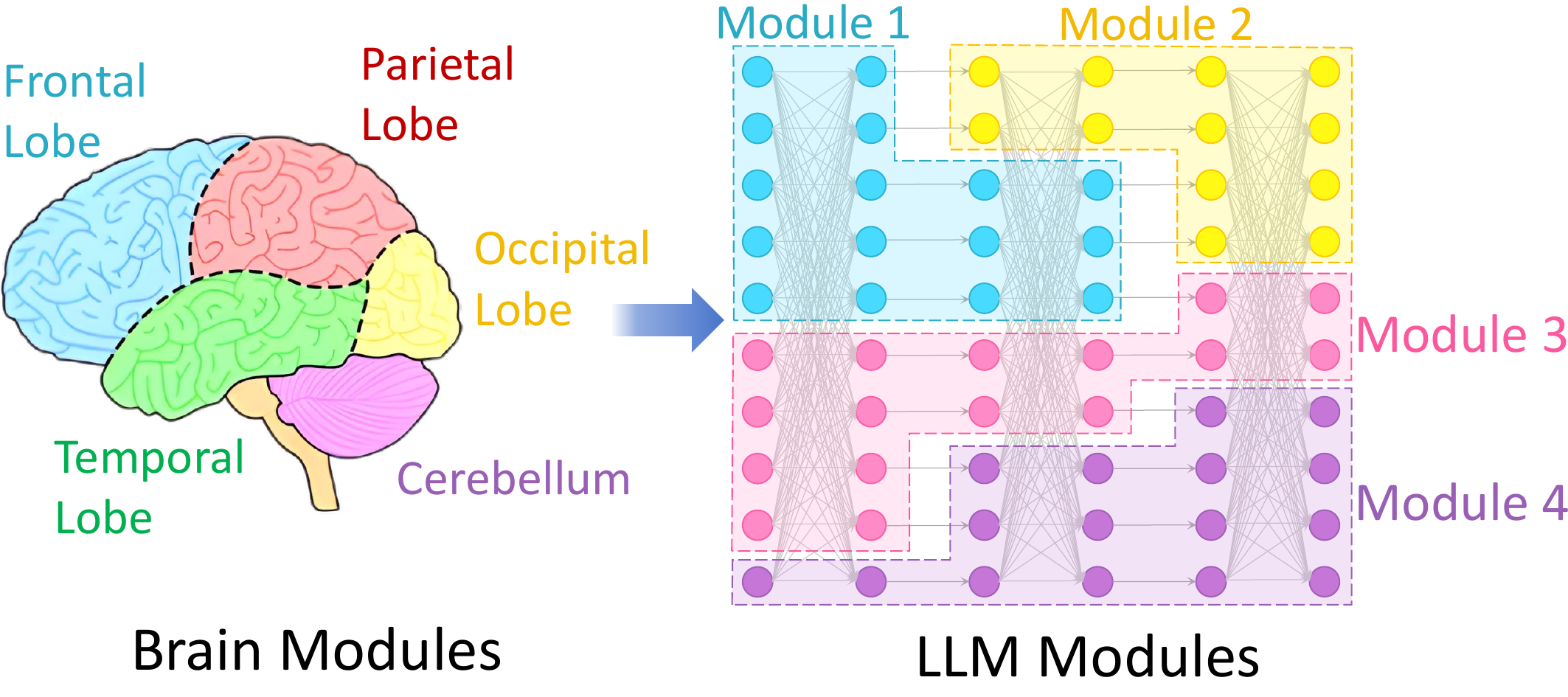}
    \caption{Illustrations of brain modules vs. LLM modules.}
    \label{fig:modules}
    \vspace{-0.2cm}
\end{figure}

Indeed, neuroscience research suggests that the human brain contains highly specialized and decoupled function modules~\cite{meunier2010modular, bullmore2009complex, kandel2013principles} shown in Fig.~\ref{fig:modules} (left).
These modules facilitate parallel processing, reduce interference between cognitive functions, and allow for greater evolutionary adaptability. 
The activation pattern differences across functions can provide meaningful insights into a similar modularity phenomenon in LLMs, as shown in Fig.~\ref{fig:modules} (right).
However, existing works only identify significantly overlapping neuron sets activated by predefined functions~\cite{xiao2024configurable}. 
How different functions are formulated by basic modules and organized within LLMs remains highly unexplored.
To our knowledge, it still lacks an effective method to discover truly decoupled function modules.
% To our knowledge, there is still a lack of an effective method to discover truly decoupled function modules.

To bridge this gap, we formulate a \textbf{function module discovery} problem, aiming to discover decoupled neuron sets that have dense co-activation performance on a specific set of samples with a shared but unknown topic. We propose an Unsupervised LLM Cross-layer MOdule Discovery (ULCMOD) framework that simultaneously disentangles the large set of neurons in the entire LLM into modules while discovering the topics of input samples related to these modules. Specifically, we formulate a dual row-column partition problem on the LLM activation matrix and introduce a novel objective function that optimizes intra-module activation density while balancing the sizes of modules. To address the combinatorially vast search space, we develop an Iterative Decoupling (IterD) algorithm that alternates between adjusting neuron and sample partitions to optimize the objective function.

We conduct extensive experiments on the Qwen2.5 LLM family. The experimental results demonstrate our method outperforms existing clustering algorithms in identifying high-quality disentangled modules. These function modules capture more meaningful semantic information and achieve superior performance in various downstream tasks. Furthermore, our qualitative analysis reveals that the discovered modules demonstrate function comprehensiveness, function hierarchy, and clear function spatial arrangement within LLMs.

Our main contributions can be summarized as follows:
\begin{enumerate}
    \item Formulate the function module discovery problem. To our knowledge, we are among the first to explore discovering decoupled function modules in LLMs, filling a critical blank in LLM interpretability research.
    \item Introduce a novel ULCMOD framework with a new objective function and an IterD algorithm, effectively identifying high-quality and structurally sound function modules.
    \item Provide extensive quantitative and qualitative analysis to validate function module discovery performance and offer insights into patterns of function organization in LLMs.
\end{enumerate}

\section{Related Work}
\label{sec:related_work}

Here, we provide related papers for a better understanding.
%The related work of this paper can be categorized into two parts: LLM Interpretability and Functional Activation Pattern Analysis.

\paragraph{LLM Interpretability}
Recently, the interpretability of LLMs has gradually drawn attention of the research community~\cite{sun2021discerning, zhao2024explainability, ji2025comprehensive, sun2025toward}, with several paradigms emerging to deconstruct their internal mechanisms, including \textit{causal tracing}, \textit{computational circuits}, and \textit{Sparse Autoencoders (SAEs)}. One paradigm, \textit{causal tracing}, seeks to understand information flow by observing how interventions on specific parameters or activations affect the final output~\cite{meng2022locating, geva2023dissecting, stolfo2023mechanistic, zhang2023towards}. These methods typically locate critical model components by modifying their states and observing the change in the final prediction. These methods could help understand the importance of different components in LLM. Another paradigm focuses on constructing \textit{computational circuits} from input to output that underlie specific model functions and behaviors, treating components like attention heads and FFN layers as fundamental units~\cite{elhage2021mathematical, wang2022interpretability, olsson2022context, hanna2023does, conmy2023towards, conmy2304towards, ameisen2025circuit}. These methods could explore the function of different attention heads or construct the entire computational circuits. More recently, methods using \textit{Sparse Autoencoders (SAEs)} have gained traction for learning more interpretable dictionary features from LLM activations~\cite{bricken2023monosemanticity, templeton2024scaling, lindsey2025biology}. Although these approaches have significantly advanced LLM interpretability, the analysis of decoupled function modules remains a largely unexplored area. This paper addresses this gap by introducing a novel framework for identifying and analyzing these modules, thereby providing a new perspective on the internal functional mechanisms of LLMs.

\paragraph{LLM Functional Activation Pattern Analysis}
Recent research of LLMs has revealed diversified activation patterns corresponding to their various functions. They analyzed activated neurons for inputs with predefined topic labels and discovered distributions among these neurons. To be specific, \citet{dai2021knowledge, meng2022locating} identified knowledge neurons responsible for expressing factual knowledge. \citet{wang2022finding, panigrahi2023task} also discovered task-specific skill neurons, which are highly predictive of task performance and whose perturbation can drastically impair the corresponding ability. \citet{gurnee2023finding, voita2023neurons} demonstrated certain neurons encode features like linguistics and positions. For multilingual processing, language-specific neurons have also been identified \cite{zhao2023unveiling, tang2024language, zhao2024large}. By providing inputs designed to elicit varied functionalities, \citet{zhang2023emergent, xiao2024configurable} observed that specific neurons are responsible for distinct functions. Using SAEs to learn dictionary features for LLM activations, \citet{templeton2024scaling} found that the embeddings of the dictionary features form clusters that align with their semantics. Previous research has identified functional neuron sets, but these sets exhibited considerable overlap~\cite{xiao2024configurable}, meaning truly decoupled modules remained undiscovered. In this paper, we discover and interpret the basic function modules, providing a clearer insight into the model's internal functional organization.

\section{Methodology}
\label{sec:methodology}

% 提出问题
To discover function modules within LLMs, we first propose an optimization problem called \textit{Neuron-Sample Dual Partitioning}.
Subsequently, we propose an iterative partitioning algorithm to solve this problem.

\subsection{Neuron-Sample Dual Partitioning Problem}
\label{sec:methodology:problem}
% 整体联合聚类的思想介绍
An LLM consists of massive hidden units that can generate intermediate scalar values for various inputs, typically referred to as neurons.
Our goal is to disentangle these neurons into several function modules, each characterized by dense co-activation on a specific group of samples that share a common topic.

\begin{figure}[t]
    \centering
    \includegraphics[width=0.75\columnwidth]{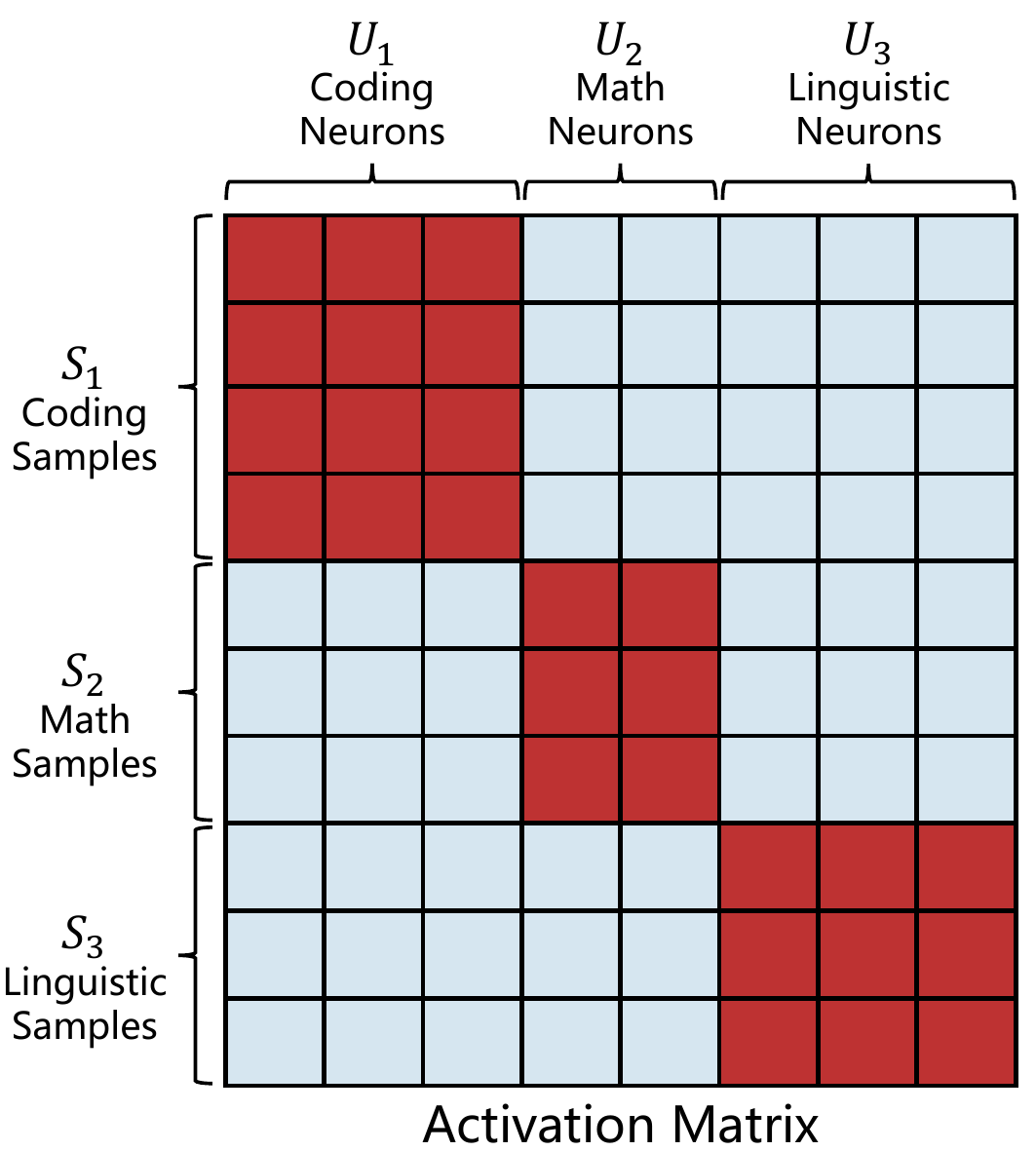}
    \caption{Illustration of function modules. Each red block represents a function module, associated with a set of neurons and a set of representative samples employing it.}
    \label{fig:method_figure1}
    \vspace{-0.1cm}
\end{figure}

% 为什么要联合聚类？ 模块功能未知 ->  需要同时发现
Specifically, we assume an LLM consists of several latent function modules associated with a set of neurons and a set of representative samples that utilize it. 
However, what functions form decoupled modules in an LLM is initially unknown, making it infeasible to identify different function modules with neurons activated by manually selected samples.
Consequently, we need to simultaneously discover both the neuron composition and the functional semantics of these decoupled modules. 
Formally, given a set of neurons $U = \{u_1, u_2, \ldots, u_N\}$ in an LLM. A set of input samples $S = \{s_1, s_2, \ldots, s_M\}$ goes through the LLM and forms an activation matrix $A \in \mathbb{R}^{N \times M}$, where $A_{n,m}$ represents the activation value of neuron $u_n$ for sample $s_m$.
We aim to find $K$ function modules $F = \{(S_1, U_1), (S_2, U_2), \ldots, (S_K, U_K)\}$, where $\mathcal{P}_S = \{S_1, S_2, \ldots, S_K\}$ and $\mathcal{P}_U = \{U_1, U_2, \ldots, U_K\}$ are the samples and neurons that belong to each functional module, which is shown in Fig.~\ref{fig:method_figure1} 

% \paragraph{}

% 问题的正式定义
To simplify the problem, we restrict each sample and neuron assigned to one and only one functional module, representing the primary function they involve.
This approach helps identify the overall semantics of functions through their core samples and neurons.
With this semantic foundation established, we can then use activation analysis to discover additional function modules that each sample and neuron participates in.

Therefore, we formalize a Neuron-Sample Dual Partitioning optimization problem, \ie, maximizing $\mathcal{L}(F)$ subject to:
\[
\begin{aligned}
% \small 
     &\textbf{(I) Completeness:} \quad \bigcup_{k=1}^K S_k = S, \quad \bigcup_{k=1}^K U_k = U, \\
 &\textbf{(II) Exclusivity:} \quad \forall \ k \neq j, \quad S_k \cap S_j = \emptyset, \quad  U_k \cap U_j = \emptyset, \\
     &\textbf{(III) Non-Emptiness:} \quad \forall \ k \in [1, K], \quad   S_k \neq \emptyset, \quad  U_k \neq \emptyset,
\end{aligned}
\]
% \[
% \begin{aligned}
% & \qquad \qquad \arg \max_{F} \quad \mathcal{L}(F) \\
% % \small
% \text{s.t.} ~ 
%     & \bigcup_{k=1}^K S_k = S, ~ \bigcup_{k=1}^K U_k = U  ~(\textbf{Completeness}), \\
% & S_k \cap S_j = \emptyset, ~ U_k \cap U_j = \emptyset, ~ \forall k \neq j  ~(\textbf{Exclusivity}), \\
%     & S_k \neq \emptyset, ~ U_k \neq \emptyset, ~ \forall k = 1, \ldots, K  ~(\textbf{Non-Emptiness}),
% \end{aligned}
% \]
where $\mathcal{L}(F)$ is the objective function optimizing the modularity of function modules $F$, which is described subsequently.

\begin{algorithm}[t!]
\caption{IterD for Function Modules Discovery}
\label{alg:algorithm}

\textbf{Input}: Activation Matrix $A$, Sample Set $S$, Neuron Set $U$

\textbf{Hyper-Parameters}: Number of Modules $K$

\textbf{Output}: $F = \{(S_k, U_k)\}_{k=1}^K$ that maximizes $\mathcal{L}(F)$

\begin{algorithmic}[1]
\STATE Initialize partition $F_0$.
\STATE $t \leftarrow 0$.
\REPEAT
    \STATE Let $\{(S_k^t, U_k^t)\}_{k=1}^K = F_t$.

    {\color{blue}{\COMMENT{\textit{Step 1: Optimize Neuron Assignments}}}}
     
    \STATE Create a temporary neuron partition \\ $\mathcal{P}'_U = \{U'_1, \dots, U'_K\} = \{U_k^t\}_{k=1}^K$.
    \FOR{each neuron $u \in U$}
        \STATE Compute $k_u^* \leftarrow \underset{k \in \{1, \ldots, K\}}{\arg\max} \, \mathcal{L}(F_{u \to k})$ based on $S_k^t$.
        \STATE Assign $u$ to its new group in $\mathcal{P}'_U$.
    \ENDFOR

    {\color{blue}{\COMMENT{\textit{Step 2: Optimize Sample Assignments}}}}
    
    \STATE Create a temporary sample partition \\ $\mathcal{P}'_S = \{S'_1, \dots, S'_K\} = \{S_k^t\}_{k=1}^K$.
    \FOR{each sample $s \in S$}
        \STATE Compute $k_s^* \leftarrow \underset{k \in \{1, \ldots, K\}}{\arg\max} \, \mathcal{L}(F_{s \to k})$ based on $U'_k$.
        \STATE Assign $s$ to its new group in $\mathcal{P}'_S$.
    \ENDFOR

    \STATE Let $F_{t+1} \leftarrow \{(S'_k, U'_k)\}_{k=1}^K$.
    % \IF{$F_{t+1}$ is identical to $F_t$}
    %     \STATE \textbf{break}
    % \ENDIF
    \STATE $t \leftarrow t + 1$.
\UNTIL{$F_{t}$ is identical to $F_{t-1}$}
\RETURN $F_{t}$
\end{algorithmic}
\vspace{-0.1cm}
\end{algorithm}

\subsection{Optimization Objectives}
\label{subsec:Optimization_Objectives}
Intuitively, an ideal partition should exhibit strong activation within each module and a relatively balanced distribution of module sizes. Therefore, $\mathcal{L}(F)$ combines two aspects: Activation Modularity $\xi(F)$ and Balance Score $B(F)$.

\paragraph{Activation Modularity $\xi(F)$}
We hypothesize that neurons and samples belonging to the same functional module should exhibit an overall higher activation magnitude.
Intuitively, a high activation magnitude implies that a neuron has a significant impact on processing a particular sample.
We therefore define an objective to maximize the average activation within modules, which we term \textit{Activation Modularity} $\xi(F)$.

Since computation in FFNs is token-wise, we first denote the activation of neuron $u_i$ for sample $s_j$ at token $t$ as $a_{i,j,t}$.
The average activation magnitude of neuron $u_i$ on sample $s_j$ is then defined as the mean of the absolute normalized activations across all tokens:
$A_{i,j} = \frac{1}{T_j} \sum_{t=1}^{T_j} |a_{i,j,t}|$, where $T_j$ is the number of tokens in sample $s_j$.
Besides, to ensure comparable activation magnitudes across neurons, each neuron's activation is normalized by \textit{z-score} over all samples.

For a modules' partition $F$ containing $K$ function modules $(S_k, U_k)$, $\xi(F)$ is the mean activation across all neuron-sample pairs $(u, s)$ that are grouped in the same module:
\begin{equation}
\xi(F) = \frac{\sum_{k=1}^K \sum_{u_i \in U_k, s_j \in S_k} A_{i,j}}{\sum_{k=1}^K |U_k| |S_k|}.
\end{equation}
A higher $\xi(F)$ indicates stronger internal activation within the modules, suggesting that it has successfully grouped neurons and samples with strong functional relations.

\paragraph{Balance Score $B(F)$} In practice, directly optimizing for $\xi(F)$ can lead to severely imbalanced module sizes. For example, it might result in one very large module and many very small ones, even if such a neuron partitioning achieves a high $\xi(F)$. To promote balanced module sizes and penalize fragmentation, we define the balance score $B(F)$:
\begin{equation}
B(F) = \frac{K}{\sum_{k=1}^K \frac{1}{|U_k| |S_k|}}.
\end{equation}
A high $B(F)$ means the module sizes are balanced.
We employ the harmonic mean here because it is particularly sensitive to small values.
Consequently, this balance score strongly penalizes partitions where one or more modules become trivially small, ensuring that all K modules are meaningful.
Finally, our objective function $\mathcal{L}(F)$ is defined as the product of the Activation Modularity and the Module Balance Coefficient: $\mathcal{L}(F) = \xi(F) \cdot B(F)$.
Our fundamental goal is to identify a high-quality partition $F$ that maximizes $\mathcal{L}(F)$, achieving both \textbf{\textit{strong internal activation}} and \textbf{\textit{a balanced distribution of module sizes}}.

\begin{figure}[t]
    \centering
    \includegraphics[width=\columnwidth]{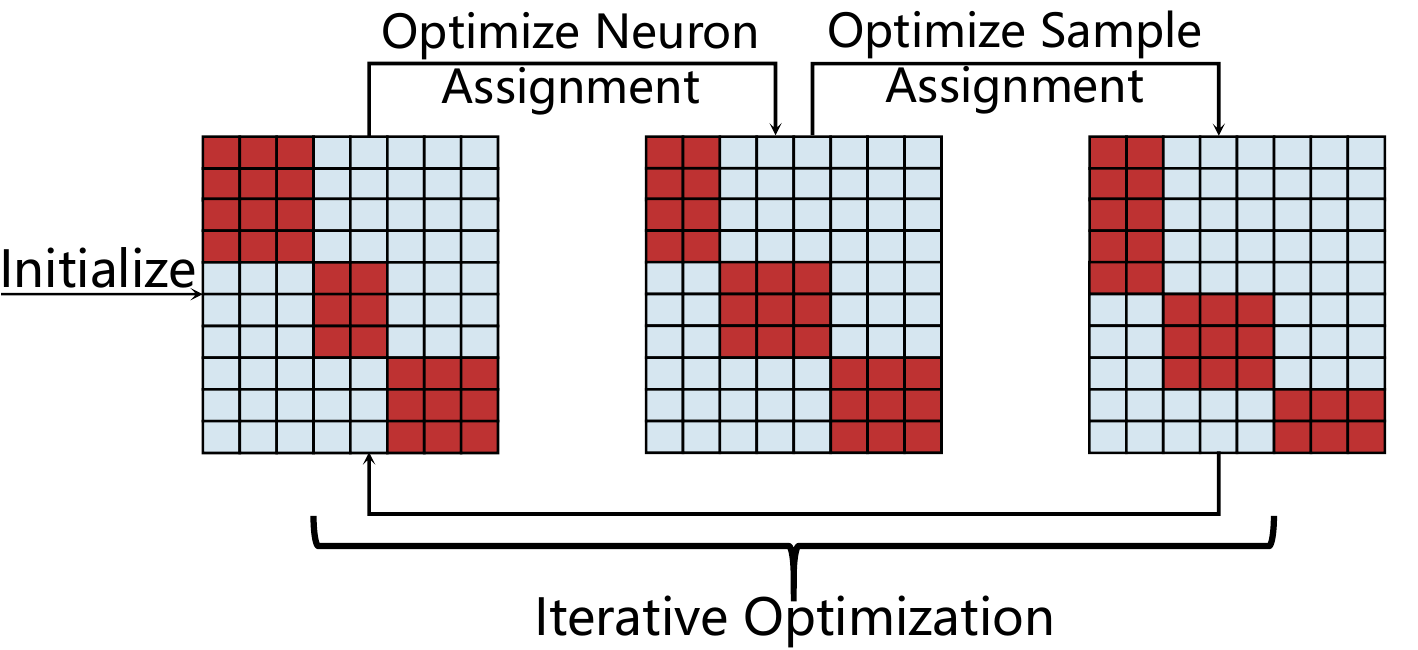} % Assuming figure2.pdf is in the same directory or a specified path
    \caption{Overview of the IterD optimization preocess.}
    \label{fig:method_figure2}
    \vspace{-0.1 cm}
\end{figure}

\subsection{Functional Region Identification Algorithm}
\label{sec:methodology:algorithm}

Directly maximizing $\mathcal{L}(F)$ is computationally challenging due to the combinatorial nature of the partitioning problem. We, therefore, propose IterD, an iterative algorithm designed to find a high-quality partition $F$.
The process, illustrated in Fig.~\ref{fig:method_figure2}, consists of an initialization stage (\textbf{Stage I}) followed by an iterative optimization stage (\textbf{Stage II}).

In \textbf{Stage I}, IterD begins with an initial partition $F_0 = \{(S_k^0, U_k^0)\}_{k=1}^K$. This can be generated by applying a basic clustering method (e.g., K-Means) to the samples and neurons to establish a reasonable starting point.

\textbf{Stage II} is iterative optimization. Given a partition $F_t = \{(S_k^t, U_k^t)\}_{k=1}^K$ at iteration $t$ (with $F_0$ from initialization), IterD iteratively refines sample and neuron assignments to greedily maximize $\mathcal{L}(F)$. Each iteration involves two steps, which will be detailed below part by part:

\paragraph{Step 1: Optimize Neuron Assignments}
In this step, the sample partition $\mathcal{P}_S^t = \{S_k^t\}_{k=1}^K$ is held fixed. IterD iterates through each neuron $u_j \in U$. For each neuron, we determine its optimal new assignment by finding the target module $k_j^*$ that greedily maximizes the objective function:
\begin{equation}
k_j^* = \underset{k \in \{1, \ldots, K\}}{\arg\max} \, \mathcal{L}(F_{j \to k}),
\end{equation}
where $F_{j \to k}$ is the resulting partition after reassigning neuron $u_j$ to module $k$. The neuron is immediately reassigned to this new module before the next neuron is considered. Therefore, the optimization for each neuron is influenced by the reassignments of all preceding neurons within the same step. After iterating through all neurons, this step yields an updated neuron partition $\mathcal{P}_U^{t+1} = \{U_k^{t+1}\}_{k=1}^K$.

\paragraph{Step 2: Optimize Sample Assignments}
Next, holding the newly updated neuron partition $\mathcal{P}_U^{t+1}$ fixed, we perform a symmetric optimization for the samples. IterD iterates through each sample $s_i \in S$. For each sample, we find the target module $k_i^*$ that greedily maximizes the objective $\mathcal{L}(F)$:
\begin{equation}
k_i^* = \underset{k \in \{1, \ldots, K\}}{\arg\max} \, \mathcal{L}(F_{i \to k}),
\end{equation}
where $F_{i \to k}$ denotes the hypothetical partition resulting from moving sample $s_i$ to module $k$. As with the neurons, the sample is immediately reassigned. This process yields the updated sample partition $\mathcal{P}_S^{t+1} = \{S_k^{t+1}\}_{k=1}^K$.
Putting these two steps together completes one full iteration, producing the new overall partition $F_{t+1} = \{(S_k^{t+1}, U_k^{t+1})\}_{k=1}^K$.

These two steps are repeated until convergence, i.e., when no neuron or sample reassignment occurs in a full iteration. More details are provided in Algo.~\ref{alg:algorithm} with a line-by-line explanation in Sec.~\ref{sec_supp:Explanation_of_IterD} of Supp.

\section{Experiments}
\label{sec:experiments}

In this section, we conducted extensive experiments to validate the effectiveness of our IterD framework (Sec.~\ref{sec:methodology:algorithm}) and empirically provide some insights.

We used three open-sourced LLMs for fair comparisons: Qwen2.5-1.5B-Instruct, Qwen2.5-3B-Instruct, and Qwen2.5-7B-Instruct. All-layer activations of these models were extracted from inference of some samples sourced from the Infinity-Instruct dataset~\cite{li2025infinity}. These samples are balancedly selected following \citet{xiao2024configurable}. More experimental details are provided in Sec.~\ref{sec_supp:Experimental_Details} of Supp.

We compare our framework against widely used clustering methods adapted for our task: \textbf{K-Means}, \textbf{Mini-Batch K-Means}~\cite{sculley2010web}, \textbf{Agglomerative Clustering}~\cite{ward1963hierarchical}, \textbf{Spectral Clustering}~\cite{ng2001spectral}, and \textbf{Spectral Co-clustering}~\cite{dhillon2001co}. We cluster the neurons with their activation vectors after PCA, \ie, row vectors $\operatorname{PCA}\left( \{A_{i,:}\}_{i\in [1, N]} \right)$. For some baselines with high complexities, we apply them on 10000 \textit{sub-cluster centroids} pre-processed by K-Means.

\begin{table*}[t!]
\centering
\resizebox{\textwidth}{!}{%
\begin{tabular}{@{}ccccc|ccc|ccc|ccc|ccc|ccc@{}}
\toprule
\multirow{2}{*}{LLM} & \multirow{2}{*}{$K$} & \multicolumn{3}{c}{K-Means} & \multicolumn{3}{c}{\makecell{Mini-Batch \\ K-Means}} & \multicolumn{3}{c}{Agglomerative} & \multicolumn{3}{c}{Spectral} & \multicolumn{3}{c}{\makecell{Spectral \\ Co-cluster}} & \multicolumn{3}{c}{\makecell{\textbf{IterD} \\ (\textbf{Ours})}} \\
\cmidrule(lr){3-5} \cmidrule(lr){6-8} \cmidrule(lr){9-11} \cmidrule(lr){12-14} \cmidrule(lr){15-17} \cmidrule(lr){18-20}
& & $\mathcal{L}(F)$ & $\xi(F)$ & $B(F)$ & $\mathcal{L}(F)$ & $\xi(F)$ & $B(F)$ & $\mathcal{L}(F)$ & $\xi(F)$ & $B(F)$ & $\mathcal{L}(F)$ & $\xi(F)$ & $B(F)$ & $\mathcal{L}(F)$ & $\xi(F)$ & $B(F)$ & $\mathcal{L}(F)$ & $\xi(F)$ & $B(F)$ \\
\midrule
\multirow{4}{*}{\makecell{Qwen2.5\\-1.5B\\-Instruct}} 
& 5  & 19.2 & 0.409 & 46.9 & 23.4 & 0.406 & 57.6 & 10.1 & 0.311 & 32.4 & 9.6  & 0.335 & 28.7 & 24.0 & 0.380 & 63.2 & \textbf{31.4} & \textbf{0.465} & \textbf{67.5}  \\
& 10 & 5.7  & 0.644 & 8.8  & 4.2  & 0.630 & 6.7  & 5.2  & 0.599 & 8.7  & 2.8  & 0.566 & 5.0  & 7.9  & 0.522 & 15.1 & \textbf{12.0} & \textbf{0.711} & \textbf{16.9}  \\
& 15 & 0.9  & 0.829 & 1.1  & 1.9  & 0.752 & 2.6  & 1.9  & 0.655 & 2.9  & 0.7  & 0.750 & 0.9  & 3.8  & 0.583 & 6.4  & \textbf{6.6}  & \textbf{0.892} & \textbf{7.5}   \\
& 20 & 0.4  & 0.943 & 0.4  & 0.5  & 0.855 & 0.6  & 1.0  & 0.731 & 1.3  & 0.3  & 0.914 & 0.3  & 2.0  & 0.636 & 3.1  & \textbf{4.4}  & \textbf{1.048} & \textbf{4.2}   \\
\midrule
\multirow{4}{*}{\makecell{Qwen2.5\\-3B\\-Instruct}} 
& 5  & 28.0 & 0.466 & 60.1 & 26.9 & 0.490 & 54.9 & 16.1 & 0.439 & 36.6 & 0.0  & 0.001 & 7.7  & 34.6 & 0.404 & 85.6 & \textbf{56.9} & \textbf{0.533} & \textbf{106.8} \\
& 10 & 12.9 & 0.747 & 17.2 & 7.1  & 0.631 & 11.3 & 7.5  & 0.624 & 12.1 & 0.6  & 0.219 & 2.6  & 12.4 & 0.570 & 21.7 & \textbf{20.8} & \textbf{0.784} & \textbf{26.5}  \\
& 15 & 1.1  & 0.853 & 1.3  & 1.4  & 0.818 & 1.7  & 2.1  & 0.836 & 2.5  & 0.4  & 0.423 & 0.8  & 1.1  & 0.559 & 1.9  & \textbf{11.3} & \textbf{0.964} & \textbf{11.8}  \\
& 20 & 0.6  & 0.974 & 0.7  & 0.7  & 0.931 & 0.7  & 0.3  & 0.940 & 0.3  & 0.3  & 0.564 & 0.5  & 0.1  & 0.659 & 0.1  & \textbf{7.3}  & \textbf{1.107} & \textbf{6.6}   \\
\midrule
\multirow{4}{*}{\makecell{Qwen2.5\\-7B\\-Instruct}} 
& 5  & 29.6 & 0.400 & 74.1 & 32.8 & 0.418 & 78.3 & 15.0 & 0.311 & 48.3 & 1.0  & 0.137 & 7.6  & 43.9 & 0.362 & 121.1& \textbf{64.6} & \textbf{0.445} & \textbf{145.2} \\
& 10 & 13.1 & 0.639 & 20.6 & 12.5 & 0.624 & 20.1 & 9.8  & 0.518 & 18.8 & 1.3  & 0.384 & 3.4  & 16.9 & 0.522 & 32.3 & \textbf{25.2} & \textbf{0.702} & \textbf{35.9}  \\
& 15 & 2.0  & 0.797 & 2.6  & 2.0  & 0.777 & 2.6  & 0.3  & 0.743 & 0.4  & 1.0  & 0.556 & 1.7  & 7.7  & 0.566 & 13.6 & \textbf{13.8} & \textbf{0.870} & \textbf{15.9}  \\
& 20 & 0.4  & 0.903 & 0.5  & 0.6  & 0.905 & 0.7  & 0.5  & 0.753 & 0.7  & 0.0  & 0.834 & 0.0  & 4.3  & 0.634 & 6.8  & \textbf{9.0}  & \textbf{1.008} & \textbf{8.9}   \\
\bottomrule
\end{tabular}%
}
\caption{Comparison of Clustering Metrics for different methods, LLMs, and number of clusters ($K$). Values for $\mathcal{L}(F)$ and $B(F)$ are scaled by $\times 10^6$.}
\label{tab:objective_comparison}
\end{table*}

\begin{figure}[t!]
    \vspace{-0.1cm}
    \centering
    \includegraphics[width=0.92\columnwidth]{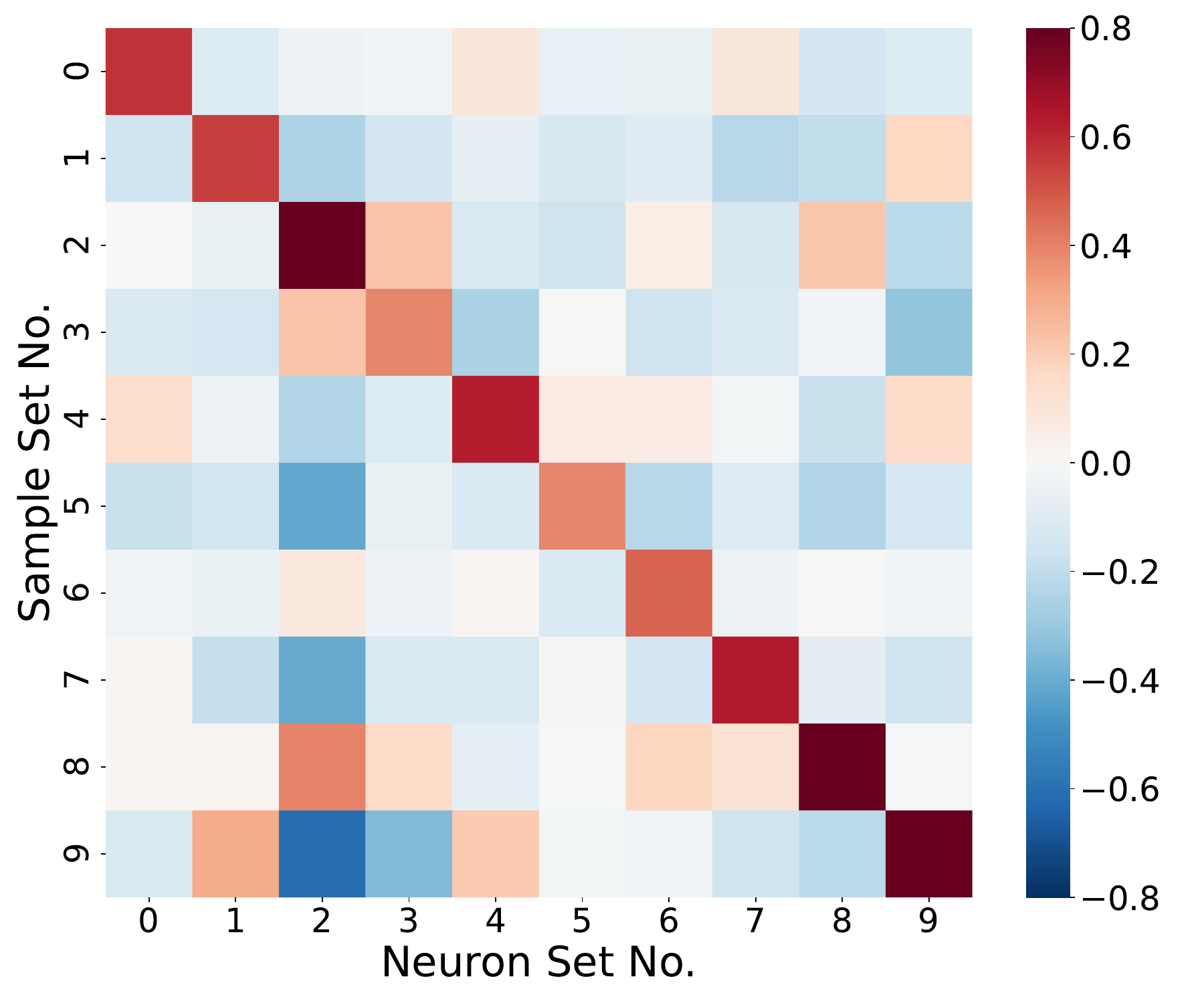}
    \caption{Average activation heatmap for discovered modules in Qwen2.5-7B-Instruct ($K=10$). Each cell $(i, j)$ shows the average activation of neuron set $U_j$ on sample set $S_i$.}
    \label{fig:block_avg_activation_heatmap}
    \vspace{-0.2cm}
\end{figure}

\begin{table*}[t]
\centering
\resizebox{0.9\textwidth}{!}{%
\begin{tabular}{@{}ccccc|cc|cc|cc|cc|cc@{}}
\toprule
\multirow{2}{*}{LLM} & \multirow{2}{*}{Cls.} & \multirow{2}{*}{$K$} & \multicolumn{2}{c}{K-Means} & \multicolumn{2}{c}{\makecell{Mini-Batch \\ K-Means}} & \multicolumn{2}{c}{Agglomerative} & \multicolumn{2}{c}{Spectral} & \multicolumn{2}{c}{\makecell{Spectral \\ Co-Clustering}} & \multicolumn{2}{c}{\makecell{\textbf{IterD} \\ (\textbf{Ours})}} \\
\cmidrule(l){4-5} \cmidrule(l){6-7} \cmidrule(l){8-9} \cmidrule(l){10-11} \cmidrule(l){12-13} \cmidrule(l){14-15}
& & & Acc & F1 & Acc & F1 & Acc & F1 & Acc & F1 & Acc & F1 & Acc & F1 \\
\midrule
\multirow{8}{*}{\makecell{Qwen2.5\\-1.5B\\-Instruct}} & \multirow{4}{*}{LR} & 5 & 0.4800 & 0.4758 & 0.5400 & 0.5404 & 0.5412 & 0.5409 & 0.5379 & 0.5300 & 0.4650 & 0.4602 & \textbf{0.5436} & \textbf{0.5428} \\
& & 10 & 0.7186 & 0.7173 & 0.6929 & 0.6927 & 0.7057 & 0.7039 & 0.7150 & 0.7150 & 0.6793 & 0.6764 & \textbf{0.7257} & \textbf{0.7259} \\
& & 15 & 0.7664 & 0.7662 & 0.7729 & 0.7724 & 0.7750 & 0.7750 & 0.7486 & 0.7474 & 0.7107 & 0.7089 & \textbf{0.8157} & \textbf{0.8158} \\
& & 20 & 0.7907 & 0.7908 & 0.7843 & 0.7839 & 0.7850 & 0.7849 & 0.7864 & 0.7862 & 0.7643 & 0.7637 & \textbf{0.8200} & \textbf{0.8200} \\
\cmidrule(l){2-15}
& \multirow{4}{*}{SVC} & 5 & 0.5414 & 0.5417 & 0.5800 & 0.5829 & 0.5921 & 0.5933 & 0.5764 & 0.5695 & 0.5343 & 0.5339 & \textbf{0.5964} & \textbf{0.5987} \\
& & 10 & 0.7307 & 0.7290 & 0.6986 & 0.6987 & 0.7479 & 0.7465 & 0.7264 & 0.7260 & 0.6929 & 0.6909 & \textbf{0.7536} & \textbf{0.7529} \\
& & 15 & 0.7771 & 0.7770 & 0.7793 & 0.7792 & 0.7843 & 0.7843 & 0.7643 & 0.7638 & 0.6950 & 0.6950 & \textbf{0.8179} & \textbf{0.8181} \\
& & 20 & 0.7986 & 0.7984 & 0.7886 & 0.7881 & 0.7843 & 0.7844 & 0.7943 & 0.7943 & 0.7671 & 0.7666 & \textbf{0.8236} & \textbf{0.8238} \\
\midrule
\multirow{8}{*}{\makecell{Qwen2.5\\-3B\\-Instruct}} & \multirow{4}{*}{LR} & 5 & 0.4700 & 0.4663 & 0.5071 & 0.5023 & 0.4864 & 0.4777 & 0.3636 & 0.3599 & 0.5007 & 0.4959 & \textbf{0.5550} & \textbf{0.5533} \\
& & 10 & 0.6907 & 0.6872 & 0.6964 & 0.6965 & 0.7264 & 0.7260 & 0.4914 & 0.4903 & 0.6436 & 0.6427 & \textbf{0.7429} & \textbf{0.7424} \\
& & 15 & 0.7421 & 0.7407 & 0.7750 & 0.7751 & 0.7664 & 0.7660 & 0.5779 & 0.5787 & 0.7093 & 0.7093 & \textbf{0.8171} & \textbf{0.8171} \\
& & 20 & 0.7879 & 0.7876 & 0.7879 & 0.7875 & 0.7886 & 0.7889 & 0.7064 & 0.7065 & 0.7621 & 0.7620 & \textbf{0.8143} & \textbf{0.8143} \\
\cmidrule(l){2-15}
& \multirow{4}{*}{SVC} & 5 & 0.5336 & 0.5354 & 0.5500 & 0.5503 & 0.5414 & 0.5442 & 0.4150 & 0.4120 & 0.5593 & 0.5598 & \textbf{0.6014} & \textbf{0.6043} \\
& & 10 & 0.7236 & 0.7207 & 0.7193 & 0.7199 & 0.7229 & 0.7225 & 0.5464 & 0.5514 & 0.6621 & 0.6623 & \textbf{0.7529} & \textbf{0.7519} \\
& & 15 & 0.7479 & 0.7467 & 0.7743 & 0.7745 & 0.7593 & 0.7586 & 0.6029 & 0.6058 & 0.6957 & 0.6959 & \textbf{0.8150} & \textbf{0.8155} \\
& & 20 & 0.7814 & 0.7815 & 0.7829 & 0.7828 & 0.7779 & 0.7779 & 0.7236 & 0.7229 & 0.7536 & 0.7531 & \textbf{0.8150} & \textbf{0.8151} \\
\midrule
\multirow{8}{*}{\makecell{Qwen2.5\\-7B\\-Instruct}} & \multirow{4}{*}{LR} & 5 & 0.5121 & 0.5086 & 0.5250 & 0.5254 & 0.5250 & 0.5194 & 0.3536 & 0.3483 & 0.4914 & 0.4886 & \textbf{0.6000} & \textbf{0.5992} \\
& & 10 & 0.7071 & 0.7049 & 0.7336 & 0.7317 & 0.7300 & 0.7291 & 0.6136 & 0.6099 & 0.6486 & 0.6462 & \textbf{0.7536} & \textbf{0.7527} \\
& & 15 & 0.7907 & 0.7900 & 0.7836 & 0.7827 & 0.7650 & 0.7634 & 0.7129 & 0.7116 & 0.7493 & 0.7484 & \textbf{0.8236} & \textbf{0.8236} \\
& & 20 & 0.7929 & 0.7924 & 0.7986 & 0.7985 & 0.7893 & 0.7886 & 0.7729 & 0.7725 & 0.7779 & 0.7774 & \textbf{0.8179} & \textbf{0.8175} \\
\cmidrule(l){2-15}
& \multirow{4}{*}{SVC} & 5 & 0.5600 & 0.5610 & 0.5693 & 0.5709 & 0.5707 & 0.5706 & 0.4107 & 0.4077 & 0.5429 & 0.5436 & \textbf{0.6293} & \textbf{0.6297} \\
& & 10 & 0.7357 & 0.7335 & 0.7586 & 0.7570 & 0.7564 & 0.7552 & 0.6536 & 0.6506 & 0.6650 & 0.6648 & \textbf{0.7750} & \textbf{0.7743} \\
& & 15 & 0.7907 & 0.7901 & 0.7750 & 0.7742 & 0.7893 & 0.7884 & 0.7364 & 0.7353 & 0.7429 & 0.7419 & \textbf{0.8207} & \textbf{0.8206} \\
& & 20 & 0.8000 & 0.7996 & 0.8036 & 0.8033 & 0.7993 & 0.7987 & 0.7607 & 0.7593 & 0.7621 & 0.7620 & \textbf{0.8214} & \textbf{0.8210} \\
\bottomrule
\end{tabular}%
}
\caption{Comparison of Test Accuracy and Macro-F1 for different methods, LLMs, classifiers, and number of clusters ($K$). Features are derived from average function module activations.}
\label{tab:test_prediction_comparison}
\vspace{-0.1cm}
\end{table*}

\begin{figure}[t!]
    \vspace{-0.1cm}
    \centering  
    \includegraphics[width=0.92\columnwidth]{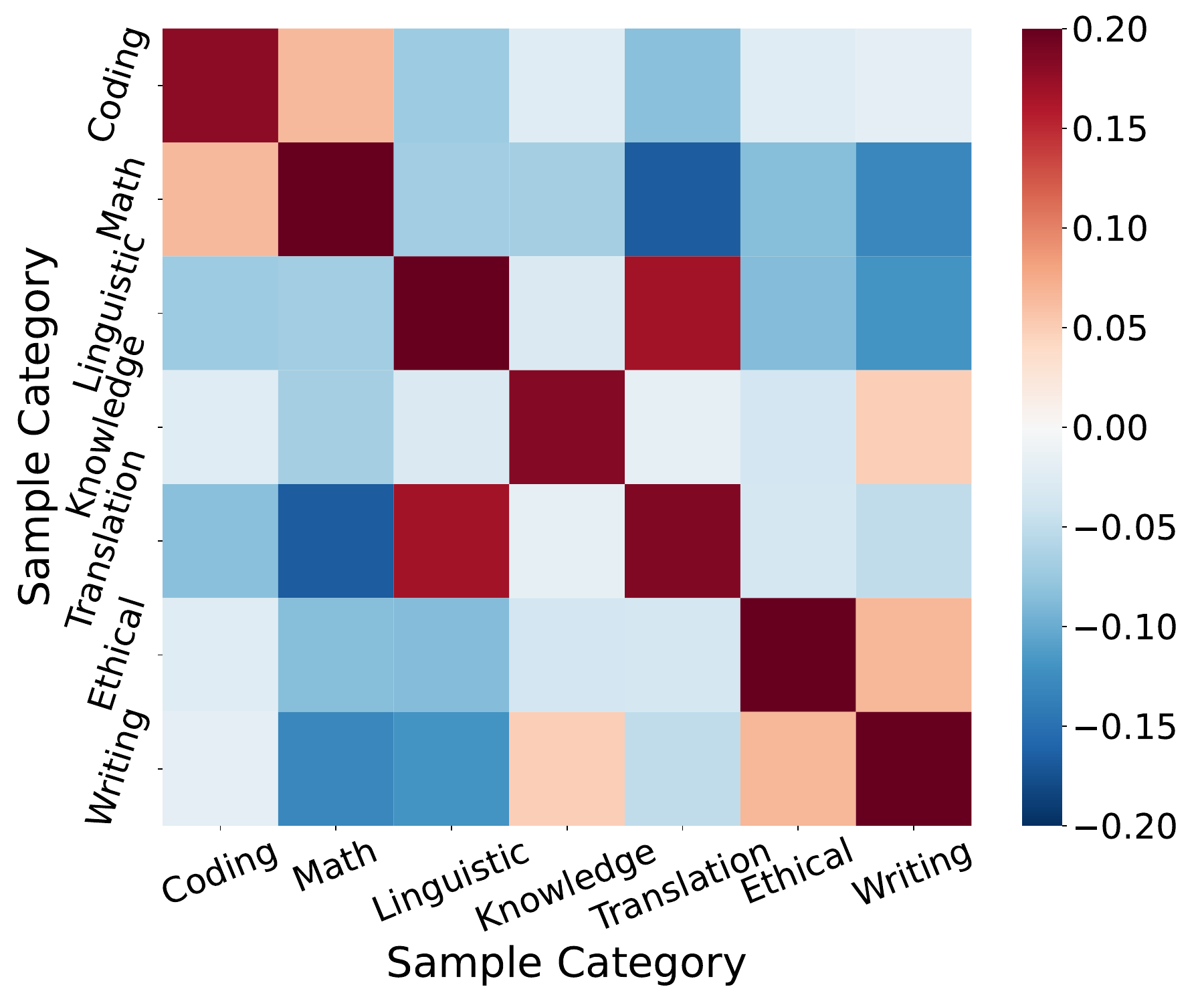}
    \caption{Visualization of sample category similarity in Qwen2.5-7B-Instruct ($K=10$). The value in cell $(i, j)$ represents the average cosine similarity between the feature vectors ($\mathbf{x}_{s}$) of samples from category $i$ and category $j$.}
    \label{fig:category_similarity}
    \vspace{-0.2cm}
\end{figure}

\subsection{Performance Comparison}
\label{sec:experiments:performance_comparison}
% 怎么引出L(F)的，动机写一下

We evaluate the performance of our framework and baselines via our optimization objective $\mathcal{L}(F)$ in Sec.~\ref{sec:methodology:problem} and a proposed novel Function Module Informativeness metric $\mathcal{I}(F)$, which comprehensively assesses the quality of the discovered function modules.

\paragraph{Comparison of $\mathcal{L}(F)$} As shown in Tab.~\ref{tab:objective_comparison}, IterD consistently and substantially outperforms all baselines in optimizing $\mathcal{L}(F)$ across all LLMs and cluster counts ($K$), stemming from both higher activation modularity ($\xi(F)$) and a more balanced size distribution ($B(F)$).
Therefore, the function modules discovered by our framework strictly match the motivation of our task.

Recall in Sec.~\ref{sec:methodology:problem}, we find $K$ function modules $F = \{(S_1, U_1), (S_2, U_2), \ldots, (S_K, U_K)\}$, where sample group $S_i$ and neuron group $U_i$ should have a clear function correspondence.
Here, we visualize such function correspondences in Fig.~\ref{fig:block_avg_activation_heatmap} for our modules discovered in Qwen2.5-7B-Instruct ($K=10$), which can be well illustrated via the \textit{co-activation patterns}, \ie, a highlighted structure in a co-activation heatmap.
From Fig.~\ref{fig:block_avg_activation_heatmap}, we can observe a clear block-diagonal highlighted structure (red), showing the discovered $S_i$ and $U_i$ are aligned functionally.
In contrast, the non-diagonal low activations (blue/white) indicate minimal interactions between non-corresponding groups $S_i$ and $U_j$ ($i\ne j$).
The strong comparison provides clear visual evidence that our algorithm successfully discovered \textit{internally co-activating} and \textit{functionally disentangled}  modules in a fully unsupervised manner.

% The high-activation values (dark red) along the diagonal confirm strong activation within each corresponding sample-neuron cluster pair $(S_k, U_k)$. Conversely, the off-diagonal elements are predominantly low-activation (blue/white), indicating minimal interaction between non-corresponding clusters. This provides clear visual evidence that our algorithm successfully isolates functionally distinct and internally co-activating modules.

\paragraph{Comparison of $\mathcal{I}(F)$}

To quantify the relevance between function modules and function labels, we propose a new metric called \textbf{Function Modular Informativeness}, denoted as $\mathcal{I}(F)$. For each sample $s_j \in S$, we construct an \textit{activation pattern vector} $\mathbf{x}_{s_j} \in \mathbb{R}^{K}$, where its $k$-th component $\mathbf{x}_{s_j,k}$ denotes how strongly $s_j$ activates the $k$-th function modules $U_k$, \ie, $\mathbf{x}_{s_j,k} = \frac{1}{|U_k|} \sum_{u_i \in U_k} A_{i,j}$.
It quantitatively describes $s_j$'s interactions with all identified modules.
Specifically, we measure $\mathcal{I}(F)$ via the generalization performance of a linear classifier $f_{\theta}$ trained to predict the function labels $y_j$ based on $\mathbf{x}_{s_j}$.
We use either Logistic Regression~\cite{cox1958regression} or a linear Support Vector Classifier (SVC)~\cite{cortes1995support}.
Formally, 
\begin{equation}
\mathcal{I}(F)=\operatorname{Acc}\left(f_{\theta^*}\right), \quad
\theta^*=\arg\min_{\theta}\frac{1}{|S|}\sum_{s_j\in S}l\bigl(f_\theta(\mathbf{x}_{s_j}),y_j\bigr),
\end{equation}
where $l(\cdot)$ is the standard classification loss, and $\operatorname{Acc}(\cdot)$ denotes the accuracy or F1 score on the unseen test samples.

% Formally, $\mathcal{I}(F)=\operatorname{Acc}\left(f_{\theta^*}\right)$, where $\theta^*=\operatorname{argmin}_{\theta} \frac{1}{|S|}\sum_{s_j\in S}l(f_\theta(\mathbf{x}_{s_j});y_j)$ ($l(\cdot)$ denotes the logistic loss) and $\operatorname{Acc\left(\cdot\right)}$ denote Accuracy or F1 score.

% feature vectors 质量更高
As shown in Tab.~\ref{tab:test_prediction_comparison}, our induced feature vectors $\mathbf{x}_s$ achieve consistently higher classification accuracies and macro-F1 scores among all model sizes and values of $K$, regardless of whether a downstream Logistic Regressor or Support Vector classifier $f_\theta$.
Since such feature vectors are derived from our discovered function modules, the superior downstream performance empirically proves that our algorithm discovers informative partitions with more meaningful functions.

Besides, to further investigate whether our neuron partition can structurally resemble ground-truth sample categories, motivated by contrastive learning, we analyzed \textit{inter-category semantic disentanglement} $\operatorname{SD}_{c,c'}$ and \textit{intra-category semantic coherence} $\operatorname{SC}_c = \operatorname{SD}_{c,c'}$ via Cos-similarities between those activation pattern vectors ($\mathbf{x}_{s}$) with:
\begin{align}
    \begin{split}
        \operatorname{SD}_{c,c'} &= \frac{1}{|S^{[c]}|\cdot |S^{[c']}|} \sum_{s\in S^{[c]}} \sum_{s' \in S^{[c']}}\operatorname{Cos} \left( x_s, x_{s'} \right),
    \end{split}
\end{align}
where $S^{[c]}$ denotes the set of all samples of category $c\in [1,C]$.
These $\{\operatorname{SD}_{c,c'}\}$ values are visualized in Fig.~\ref{fig:category_similarity}, which reveals two findings:
(1) Same-category samples (\eg, \textit{Math} and \textit{Coding}) exhibit highly similar activation patterns, whereas distinct-category samples display contrasting phenomena.
(2) Some non-ignorable inter-category semantic connections can be successfully identified via those (corresponding) relatively high non-diagonal co-activations.
For example, the notable semantic similarity $\operatorname{SD}_{c,c'}$ between categories \textit{Linguistic} and \textit{Translation} achieves supervising $0.168$, indicating reasonably similar behaviors of LLMs when dealing with such highly related tasks.

\subsection{Empirical Insights of Function Modules} 
\label{subsec:Empirical_Insights_of_Function_Modules}

\begin{figure}[t]
    \vspace{-0.375cm}
    \centering
    \begin{subfigure}[b]{0.465\textwidth}
        \centering
        \includegraphics[width=\textwidth]{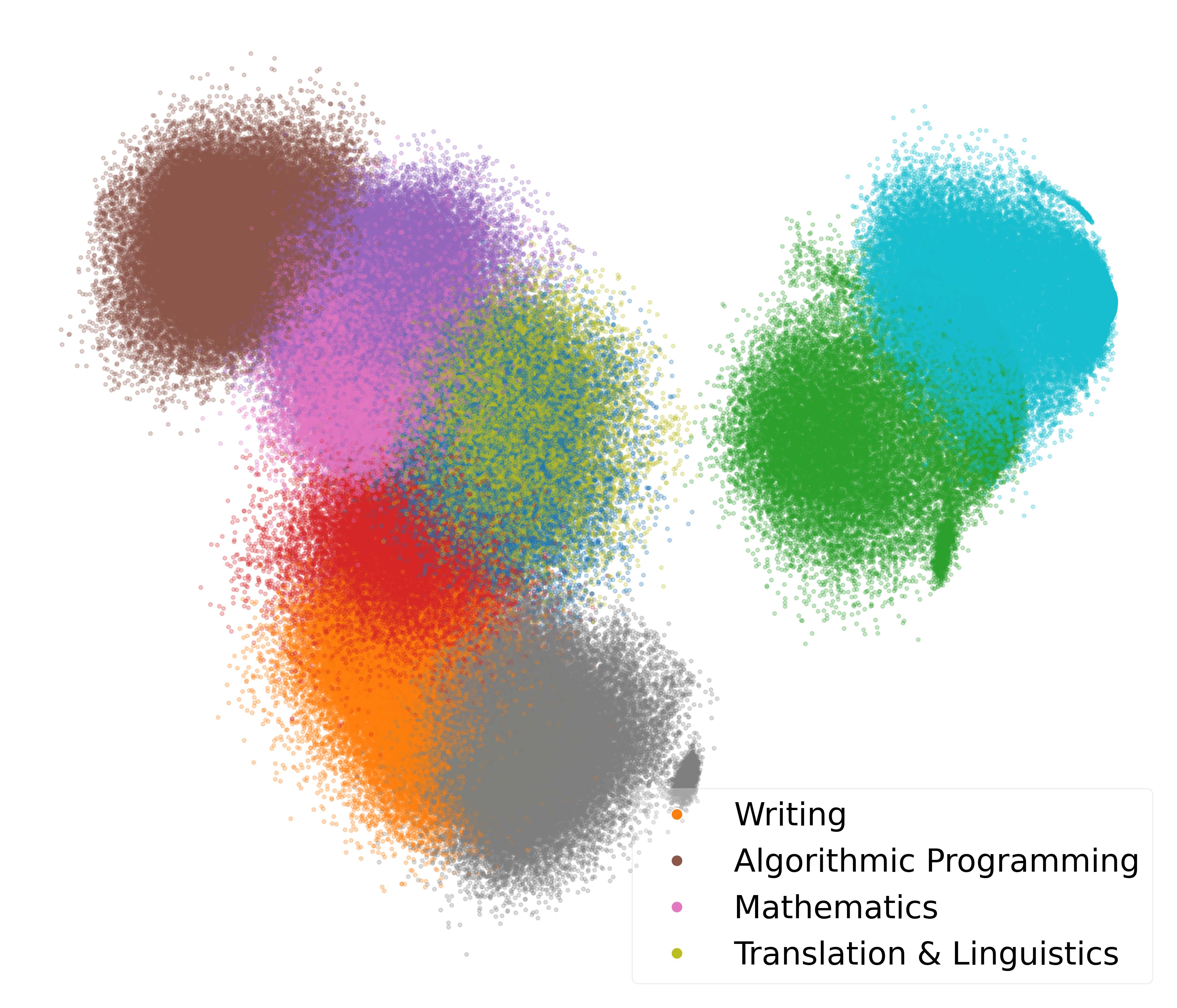}
        \caption{Function modules for $K=10$}
        \vspace{-0.1cm}
        \label{fig:vis_k10}
    \end{subfigure}
    \hfill %
    \begin{subfigure}[b]{0.465\textwidth}
        \centering
        \includegraphics[width=\textwidth]{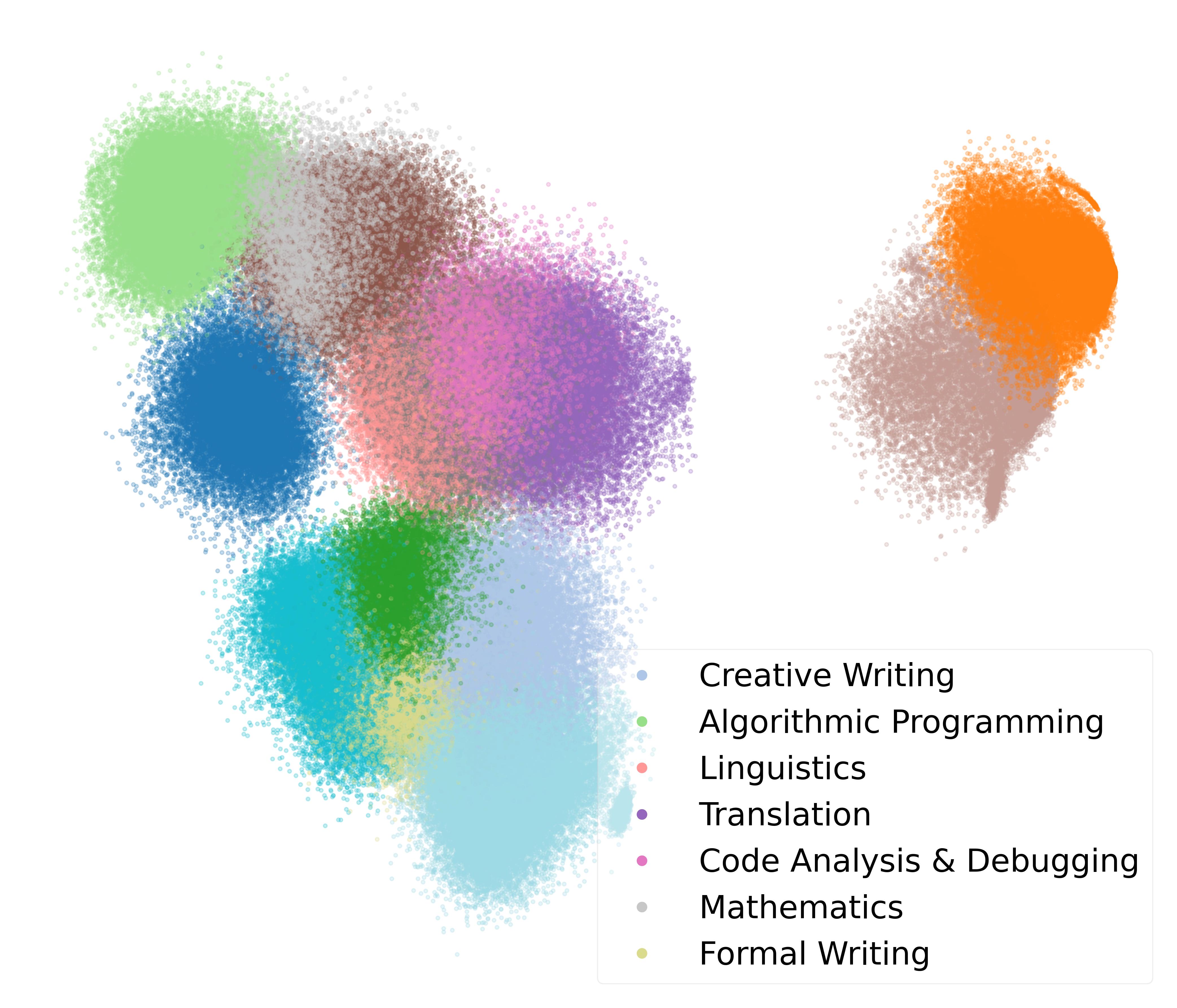}
        \caption{Function modules for $K=15$}
        \vspace{-0.1cm}
        \label{fig:vis_k15}
    \end{subfigure}
    \hfill %
    \begin{subfigure}[b]{0.465\textwidth}
        \centering
        \includegraphics[width=\textwidth]{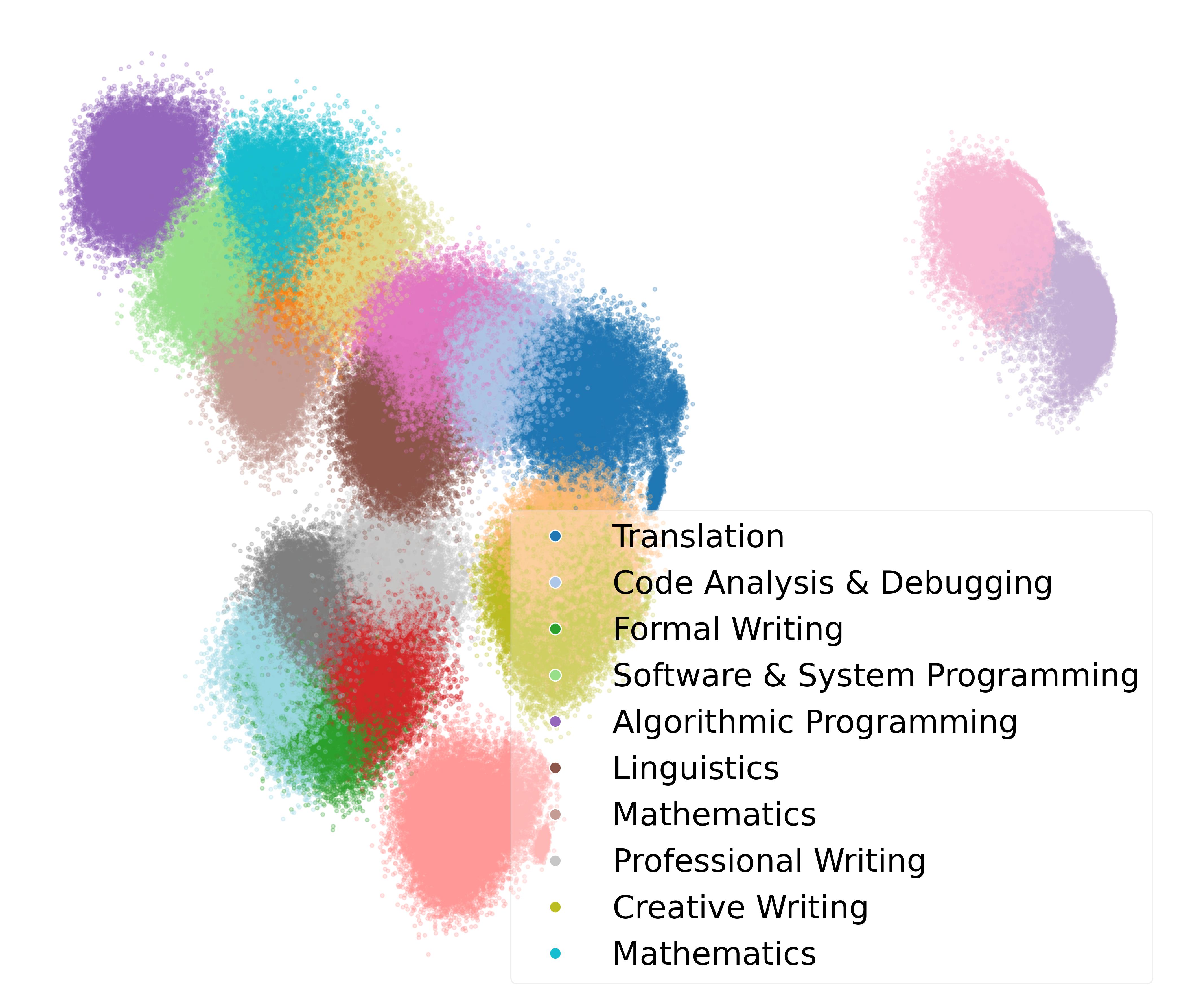}
        \caption{Function modules for $K=20$}
        \vspace{-0.1cm}
        \label{fig:vis_k20}
    \end{subfigure}
    \caption{Visualization of discovered function modules.% with variable $K$s in Qwen2.5-3B-Instruct. % Each point represents a neuron, colored by its assigned function module.
    }
    \label{fig:visualizations}
\end{figure}

To provide qualitative insights into the nature of our identified function modules, we visualized the neuron groups for Qwen2.5-3B-Instruct with varying module numbers (\ie, $K \in \{10, 15, 20\}$) in Fig.~\ref{fig:visualizations}, where each point represents a neuron and colors denote different function modules.

\begin{figure}
    \centering
    \includegraphics[width=\linewidth]{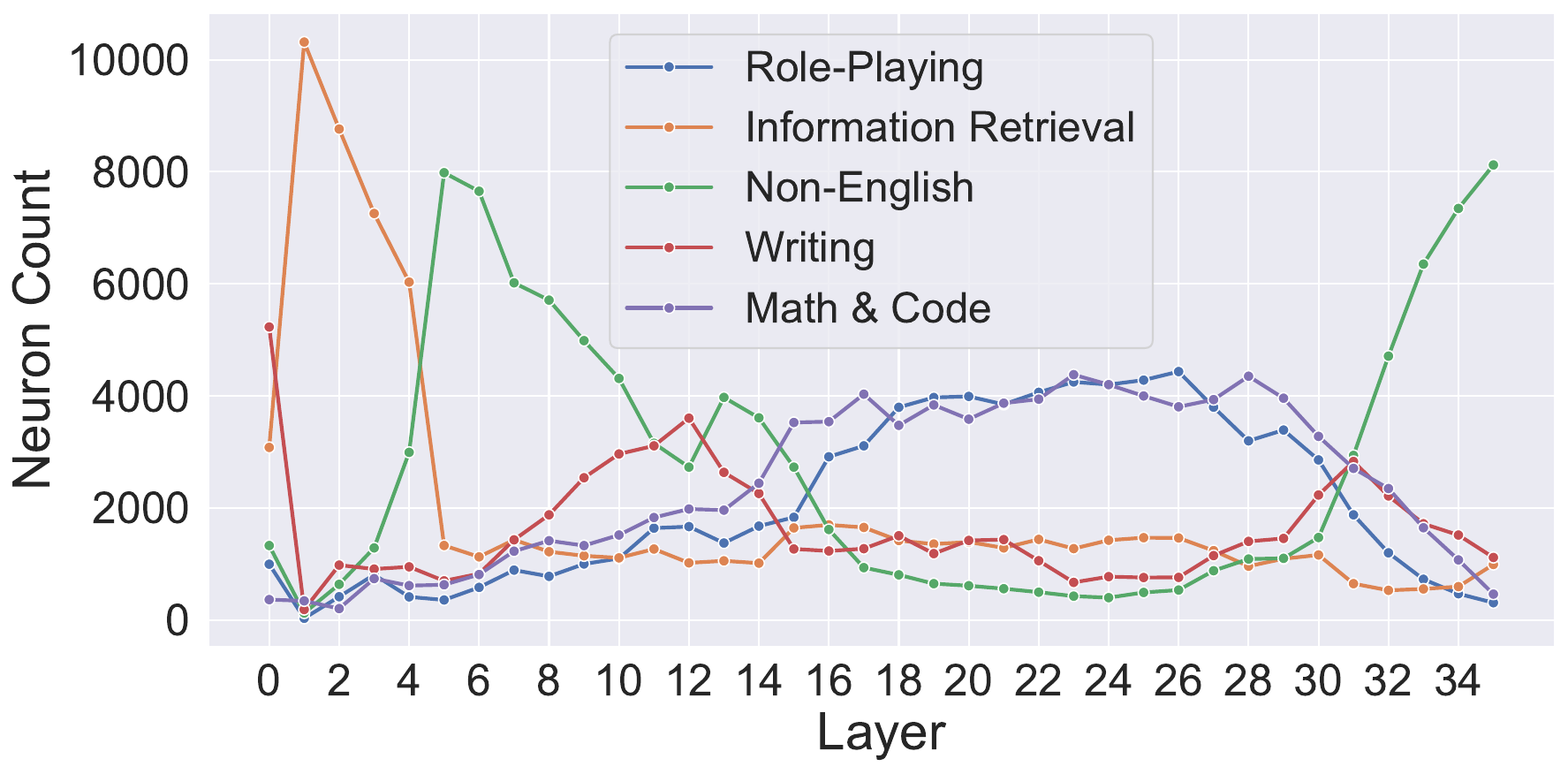}
    \caption{The studies of how neuron count is distributed in different layers inside each function module for Qwen2.5-3B-Instruct ($K = 5$).}
    \label{fig:layer_dist_combined}
    % \vspace{-0.2cm}
\end{figure}

To further functionally analyze these modules, we first determine their specific functions from the LLMs' summaries for some highly activated samples.
We provide the details in Sec.~\ref{sec_supp:Function_Determination_via_LLM} of Supp.
For example, we inferred the function corresponding to the brown module should be \textit{Algorithmic Programming}.
The respective functions of other modules are labeled in Fig.~\ref{fig:visualizations}. Besides, to study how LLMs perform or realize a function layer by layer, we illustrate in Fig.~\ref{fig:layer_dist_combined} how the neuron count of a function module is distributed in different layers.

Based on these figures, we observe several key findings:
\paragraph{Comprehensive Function Discovery} All discovered modules correspond to interpretable functions tightly aligned with human knowledge or skill.
Moreover, some critical functions consistently emerge across all levels of granularity $K$ (such as Programming, Mathematics, and Writing), demonstrating the merits of our module partition algorithm.

\paragraph{Function Hierarchy}
Comparing the function visualizations for different values of $K$ provides a clear hierarchical organization of such functions.
A small $K$ presents coarse-grained function disentanglement, while a large $K$ gives a fine-grained function discovery akin to the "\textit{zoom in}" effects.
The examples described below evidently verify this insight:
\begin{itemize}
    \item \textbf{Programming:} A large module \textit{Algorithmic Programming} at $K=10$ is split into several more specialized modules at $K=20$ (\eg, \textit{Code Analysis \& Debugging}, and \textit{Software \& System Programming}), showing a clear divergence of engineering pathways.

    \item \textbf{Writing:} The general \textit{Writing} module from $K=10$ is refined into \textit{Creative Writing}, \textit{Formal Writing}, and \textit{Professional Writing} modules at $K=20$, demonstrating an essential neuron hierarchy inside LLMs naturally supporting different writing styles.

\end{itemize}

\paragraph{Insights of Function Spatial Arrangement} The locality of these modules is well aligned with their corresponding semantical functions.
In other words, semantically related functions are located closely, revealing an intrinsic \textit{connected} structure of related skills.
For instance, functions related to programming, math, and science always appear nearby, forming a large "Techniques" connected component.
In contrast, distant clusters often show low semantic relevance.
Furthermore, a central hub of complex cognitive tasks (\eg, \textit{Linguistics} and \textit{Translation}) often emerges at the intersection of other modules, suggesting that they are core capabilities leveraged across multiple domains.

\paragraph{Layer Distribution of Functions}
From Fig.~\ref{fig:layer_dist_combined}, different layers contribute quite variably to a function, which depends on the function's cognitive complexity. For instance, \textit{Information Retrieval} requires only lower-layer processing, whereas functions like \textit{Math \& Code} require deeper-layer analysis.

\section{Conclusion}
\label{sec:conclusion}

% This paper introduces the ULCMOD framework to find decoupled function modules in LLMs. The proposed IterD algorithm successfully identifies high-quality, disentangled modules. The analysis reveals that these function modules exhibit function comprehensiveness, function hierarchy, and clear function spatial arrangement within LLMs.

In this paper, we propose a novel framework for discovering decoupled function modules in Large Language Models, addressing the challenge that prior work often identifies functionally overlapping neuron sets.
We propose IterD, an iterative optimization algorithm, which maximizes a new objective function that prioritizes \textit{high internally co-activations} and \textit{clear functionally disentanglement} simultaneously in a fully unsupervised manner.
% We proposed IterD, an iterative optimization algorithm that co-clusters neurons and input samples by maximizing a new objective function that balances high intra-module activation density with a sound distribution of module sizes.
%
% Intuitively, an ideal partition should exhibit strong activation within each module and a relatively balanced distribution of module sizes.
%
% The strong comparison provides clear visual evidence that our algorithm successfully discovered \textit{internally co-activating} and \textit{functionally disentangled}  modules in a fully unsupervised manner.
%
On the Qwen2.5 LLMs family, our IterD framework significantly outperforms standard baselines.
% Further experimental analysis reveals that the discovered function modules exhibit function comprehensiveness, function hierarchy, and clear function spatial arrangement within LLMs.
% Crucially, features derived from the discovered modules yield superior performance in downstream tasks, confirming their functional significance. 
% Our work provides a new analysis framework for inherent functional architectures and mechanisms of LLMs, paving the way for more flexible, effective, and trustworthy model analysis, intervention, and improvement.
Moreover, our qualitative analysis reveals that the discovered modules show function comprehensiveness, function hierarchy, and clear function spatial arrangement within LLMs.
Our work provides a novel tool for interpreting LLMs' function modules, filling a critical gap in LLMs' interpretability research.

% {\scriptsize
% QWERTYUIOPASDFGHJKLZXCVBNM

% qwertyuiopasdfghjklzxcvbnm
% }

\clearpage

\section*{Acknowledgements}

This work is partly supported by the National Natural Science Foundation of China (No. 62306255, 92370204), the National Key Research and Development Program of China (No. 2023YFF0725000), the Guangdong Basic and Applied Basic Research Foundation (No. 2024A1515011839), and the Education Bureau of Guangzhou Municipality.
\bibliography{main}

\begin{thebibliography}{48}
\providecommand{\natexlab}[1]{#1}

\bibitem[{Ameisen et~al.(2025)Ameisen, Lindsey, Pearce, Gurnee, Turner, Chen, Citro, Abrahams, Carter, Hosmer, Marcus, Sklar, Templeton, Bricken, McDougall, Cunningham, Henighan, Jermyn, Jones, Persic, Qi, Ben~Thompson, Zimmerman, Rivoire, Conerly, Olah, and Batson}]{ameisen2025circuit}
Ameisen, E.; Lindsey, J.; Pearce, A.; Gurnee, W.; Turner, N.~L.; Chen, B.; Citro, C.; Abrahams, D.; Carter, S.; Hosmer, B.; Marcus, J.; Sklar, M.; Templeton, A.; Bricken, T.; McDougall, C.; Cunningham, H.; Henighan, T.; Jermyn, A.; Jones, A.; Persic, A.; Qi, Z.; Ben~Thompson, T.; Zimmerman, S.; Rivoire, K.; Conerly, T.; Olah, C.; and Batson, J. 2025.
\newblock Circuit Tracing: Revealing Computational Graphs in Language Models.
\newblock \emph{Transformer Circuits Thread}.

\bibitem[{Bills et~al.(2023)Bills, Cammarata, Mossing, Tillman, Gao, Goh, Sutskever, Leike, Wu, and Saunders}]{bills2023language}
Bills, S.; Cammarata, N.; Mossing, D.; Tillman, H.; Gao, L.; Goh, G.; Sutskever, I.; Leike, J.; Wu, J.; and Saunders, W. 2023.
\newblock Language models can explain neurons in language models.
\newblock \url{https://openaipublic.blob.core.windows.net/neuron-explainer/paper/index.html}.

\bibitem[{Bricken et~al.(2023)Bricken, Templeton, Batson, Chen, Jermyn, Conerly, Turner, Anil, Denison, Askell, Lasenby, Wu, Kravec, Schiefer, Maxwell, Joseph, Hatfield-Dodds, Tamkin, Nguyen, McLean, Burke, Hume, Carter, Henighan, and Olah}]{bricken2023monosemanticity}
Bricken, T.; Templeton, A.; Batson, J.; Chen, B.; Jermyn, A.; Conerly, T.; Turner, N.; Anil, C.; Denison, C.; Askell, A.; Lasenby, R.; Wu, Y.; Kravec, S.; Schiefer, N.; Maxwell, T.; Joseph, N.; Hatfield-Dodds, Z.; Tamkin, A.; Nguyen, K.; McLean, B.; Burke, J.~E.; Hume, T.; Carter, S.; Henighan, T.; and Olah, C. 2023.
\newblock Towards Monosemanticity: Decomposing Language Models With Dictionary Learning.
\newblock \emph{Transformer Circuits Thread}.
\newblock Https://transformer-circuits.pub/2023/monosemantic-features/index.html.

\bibitem[{Bullmore and Sporns(2009)}]{bullmore2009complex}
Bullmore, E.; and Sporns, O. 2009.
\newblock Complex brain networks: graph theoretical analysis of structural and functional systems.
\newblock \emph{Nature reviews neuroscience}, 10(3): 186--198.

\bibitem[{Conmy et~al.(2023{\natexlab{a}})Conmy, Mavor-Parker, Lynch, Heimersheim, and Garriga-Alonso}]{conmy2023towards}
Conmy, A.; Mavor-Parker, A.; Lynch, A.; Heimersheim, S.; and Garriga-Alonso, A. 2023{\natexlab{a}}.
\newblock Towards automated circuit discovery for mechanistic interpretability.
\newblock \emph{Advances in Neural Information Processing Systems}, 36: 16318--16352.

\bibitem[{Conmy et~al.(2023{\natexlab{b}})Conmy, Mavor-Parker, Lynch, Heimersheim, and Garriga-Alonso}]{conmy2304towards}
Conmy, A.; Mavor-Parker, A.~N.; Lynch, A.; Heimersheim, S.; and Garriga-Alonso, A. 2023{\natexlab{b}}.
\newblock Towards automated circuit discovery for mechanistic interpretability.
\newblock \emph{URL https://arxiv. org/abs/2304.14997}, 2.

\bibitem[{Cortes and Vapnik(1995)}]{cortes1995support}
Cortes, C.; and Vapnik, V. 1995.
\newblock Support-vector networks.
\newblock \emph{Machine learning}, 20: 273--297.

\bibitem[{Cox(1958)}]{cox1958regression}
Cox, D.~R. 1958.
\newblock The regression analysis of binary sequences.
\newblock \emph{Journal of the Royal Statistical Society Series B: Statistical Methodology}, 20(2): 215--232.

\bibitem[{Dai et~al.(2021)Dai, Dong, Hao, Sui, Chang, and Wei}]{dai2021knowledge}
Dai, D.; Dong, L.; Hao, Y.; Sui, Z.; Chang, B.; and Wei, F. 2021.
\newblock Knowledge neurons in pretrained transformers.
\newblock \emph{arXiv preprint arXiv:2104.08696}.

\bibitem[{Dhillon(2001)}]{dhillon2001co}
Dhillon, I.~S. 2001.
\newblock Co-clustering documents and words using bipartite spectral graph partitioning.
\newblock In \emph{Proceedings of the seventh ACM SIGKDD international conference on Knowledge discovery and data mining}, 269--274.

\bibitem[{Elhage et~al.(2021)Elhage, Nanda, Olsson, Henighan, Joseph, Mann, Askell, Bai, Chen, Conerly, DasSarma, Drain, Ganguli, Hatfield-Dodds, Hernandez, Jones, Kernion, Lovitt, Ndousse, Amodei, Brown, Clark, Kaplan, McCandlish, and Olah}]{elhage2021mathematical}
Elhage, N.; Nanda, N.; Olsson, C.; Henighan, T.; Joseph, N.; Mann, B.; Askell, A.; Bai, Y.; Chen, A.; Conerly, T.; DasSarma, N.; Drain, D.; Ganguli, D.; Hatfield-Dodds, Z.; Hernandez, D.; Jones, A.; Kernion, J.; Lovitt, L.; Ndousse, K.; Amodei, D.; Brown, T.; Clark, J.; Kaplan, J.; McCandlish, S.; and Olah, C. 2021.
\newblock A Mathematical Framework for Transformer Circuits.
\newblock \emph{Transformer Circuits Thread}.
\newblock Https://transformer-circuits.pub/2021/framework/index.html.

\bibitem[{Geva et~al.(2023)Geva, Bastings, Filippova, and Globerson}]{geva2023dissecting}
Geva, M.; Bastings, J.; Filippova, K.; and Globerson, A. 2023.
\newblock Dissecting recall of factual associations in auto-regressive language models.
\newblock \emph{arXiv preprint arXiv:2304.14767}.

\bibitem[{Geva et~al.(2020)Geva, Schuster, Berant, and Levy}]{geva2020transformer}
Geva, M.; Schuster, R.; Berant, J.; and Levy, O. 2020.
\newblock Transformer feed-forward layers are key-value memories.
\newblock \emph{arXiv preprint arXiv:2012.14913}.

\bibitem[{Gong and Sun(2024)}]{gong2024graph}
Gong, Z.; and Sun, Y. 2024.
\newblock Graph reasoning enhanced language models for text-to-sql.
\newblock In \emph{Proceedings of the 47th International ACM SIGIR Conference on Research and Development in Information Retrieval}, 2447--2451.

\bibitem[{Guo et~al.(2025)Guo, Guo, Zhu, and Sun}]{guo2025towards}
Guo, Y.; Guo, S.; Zhu, H.; and Sun, Y. 2025.
\newblock Towards Lifelong Model Editing via Simulating Ideal Editor.
\newblock In \emph{Proceedings of the 42th International Conference on Machine Learning}.

\bibitem[{Gurnee et~al.(2023)Gurnee, Nanda, Pauly, Harvey, Troitskii, and Bertsimas}]{gurnee2023finding}
Gurnee, W.; Nanda, N.; Pauly, M.; Harvey, K.; Troitskii, D.; and Bertsimas, D. 2023.
\newblock Finding neurons in a haystack: Case studies with sparse probing.
\newblock \emph{arXiv preprint arXiv:2305.01610}.

\bibitem[{Hanna, Liu, and Variengien(2023)}]{hanna2023does}
Hanna, M.; Liu, O.; and Variengien, A. 2023.
\newblock How does GPT-2 compute greater-than?: Interpreting mathematical abilities in a pre-trained language model.
\newblock \emph{Advances in Neural Information Processing Systems}, 36: 76033--76060.

\bibitem[{Huang et~al.(2025)Huang, Yu, Ma, Zhong, Feng, Wang, Chen, Peng, Feng, Qin et~al.}]{huang2025survey}
Huang, L.; Yu, W.; Ma, W.; Zhong, W.; Feng, Z.; Wang, H.; Chen, Q.; Peng, W.; Feng, X.; Qin, B.; et~al. 2025.
\newblock A survey on hallucination in large language models: Principles, taxonomy, challenges, and open questions.
\newblock \emph{ACM Transactions on Information Systems}, 43(2): 1--55.

\bibitem[{Ji et~al.(2025)Ji, Sun, Zhang, Wang, Zhuang, Gong, Shen, Qin, Zhu, and Xiong}]{ji2025comprehensive}
Ji, Y.; Sun, Y.; Zhang, Y.; Wang, Z.; Zhuang, Y.; Gong, Z.; Shen, D.; Qin, C.; Zhu, H.; and Xiong, H. 2025.
\newblock A comprehensive survey on self-interpretable neural networks.
\newblock \emph{arXiv preprint arXiv:2501.15638}.

\bibitem[{Kandel et~al.(2013)Kandel, Schwartz, Jessell, Siegelbaum, and Hudspeth}]{kandel2013principles}
Kandel, E.~R.; Schwartz, J.~H.; Jessell, T.; Siegelbaum, S.~A.; and Hudspeth, A. 2013.
\newblock Principles of neural science.

\bibitem[{Li et~al.(2025)Li, Du, Zhao, Zhang, Wang, Gao, Liu, and Lin}]{li2025infinity}
Li, J.; Du, L.; Zhao, H.; Zhang, B.-w.; Wang, L.; Gao, B.; Liu, G.; and Lin, Y. 2025.
\newblock Infinity Instruct: Scaling Instruction Selection and Synthesis to Enhance Language Models.
\newblock \emph{arXiv preprint arXiv:2506.11116}.

\bibitem[{Lindsey et~al.(2025)Lindsey, Gurnee, Ameisen, Chen, Pearce, Turner, Citro, Abrahams, Carter, Hosmer, Marcus, Sklar, Templeton, Bricken, McDougall, Cunningham, Henighan, Jermyn, Jones, Persic, Qi, Thompson, Zimmerman, Rivoire, Conerly, Olah, and Batson}]{lindsey2025biology}
Lindsey, J.; Gurnee, W.; Ameisen, E.; Chen, B.; Pearce, A.; Turner, N.~L.; Citro, C.; Abrahams, D.; Carter, S.; Hosmer, B.; Marcus, J.; Sklar, M.; Templeton, A.; Bricken, T.; McDougall, C.; Cunningham, H.; Henighan, T.; Jermyn, A.; Jones, A.; Persic, A.; Qi, Z.; Thompson, T.~B.; Zimmerman, S.; Rivoire, K.; Conerly, T.; Olah, C.; and Batson, J. 2025.
\newblock On the Biology of a Large Language Model.
\newblock \emph{Transformer Circuits Thread}.

\bibitem[{Meng et~al.(2022)Meng, Bau, Andonian, and Belinkov}]{meng2022locating}
Meng, K.; Bau, D.; Andonian, A.; and Belinkov, Y. 2022.
\newblock Locating and editing factual associations in gpt.
\newblock \emph{Advances in neural information processing systems}, 35: 17359--17372.

\bibitem[{Meunier, Lambiotte, and Bullmore(2010)}]{meunier2010modular}
Meunier, D.; Lambiotte, R.; and Bullmore, E.~T. 2010.
\newblock Modular and hierarchically modular organization of brain networks.
\newblock \emph{Frontiers in neuroscience}, 4: 200.

\bibitem[{Ng, Jordan, and Weiss(2001)}]{ng2001spectral}
Ng, A.; Jordan, M.; and Weiss, Y. 2001.
\newblock On spectral clustering: Analysis and an algorithm.
\newblock \emph{Advances in neural information processing systems}, 14.

\bibitem[{Olsson et~al.(2022)Olsson, Elhage, Nanda, Joseph, DasSarma, Henighan, Mann, Askell, Bai, Chen, Conerly, Drain, Ganguli, Hatfield-Dodds, Hernandez, Johnston, Jones, Kernion, Lovitt, Ndousse, Amodei, Brown, Clark, Kaplan, McCandlish, and Olah}]{olsson2022context}
Olsson, C.; Elhage, N.; Nanda, N.; Joseph, N.; DasSarma, N.; Henighan, T.; Mann, B.; Askell, A.; Bai, Y.; Chen, A.; Conerly, T.; Drain, D.; Ganguli, D.; Hatfield-Dodds, Z.; Hernandez, D.; Johnston, S.; Jones, A.; Kernion, J.; Lovitt, L.; Ndousse, K.; Amodei, D.; Brown, T.; Clark, J.; Kaplan, J.; McCandlish, S.; and Olah, C. 2022.
\newblock In-context Learning and Induction Heads.
\newblock \emph{Transformer Circuits Thread}.
\newblock Https://transformer-circuits.pub/2022/in-context-learning-and-induction-heads/index.html.

\bibitem[{Panigrahi et~al.(2023)Panigrahi, Saunshi, Zhao, and Arora}]{panigrahi2023task}
Panigrahi, A.; Saunshi, N.; Zhao, H.; and Arora, S. 2023.
\newblock Task-specific skill localization in fine-tuned language models.
\newblock In \emph{International Conference on Machine Learning}, 27011--27033. PMLR.

\bibitem[{Sculley(2010)}]{sculley2010web}
Sculley, D. 2010.
\newblock Web-scale k-means clustering.
\newblock In \emph{Proceedings of the 19th international conference on World wide web}, 1177--1178.

\bibitem[{Stolfo, Belinkov, and Sachan(2023)}]{stolfo2023mechanistic}
Stolfo, A.; Belinkov, Y.; and Sachan, M. 2023.
\newblock A mechanistic interpretation of arithmetic reasoning in language models using causal mediation analysis.
\newblock \emph{arXiv preprint arXiv:2305.15054}.

\bibitem[{Sun et~al.(2021)Sun, Zhu, Qin, Zhuang, He, and Xiong}]{sun2021discerning}
Sun, Y.; Zhu, H.; Qin, C.; Zhuang, F.; He, Q.; and Xiong, H. 2021.
\newblock Discerning decision-making process of deep neural networks with hierarchical voting transformation.
\newblock \emph{Advances in Neural Information Processing Systems}, 34: 17221--17234.

\bibitem[{Sun, Zhu, and Xiong(2025)}]{sun2025toward}
Sun, Y.; Zhu, H.; and Xiong, H. 2025.
\newblock Toward Faithful Neural Network Intrinsic Interpretation With Shapley Additive Self-Attribution.
\newblock \emph{IEEE Transactions on Neural Networks and Learning Systems}.

\bibitem[{Tang et~al.(2024)Tang, Luo, Huang, Zhang, Wang, Zhao, Wei, and Wen}]{tang2024language}
Tang, T.; Luo, W.; Huang, H.; Zhang, D.; Wang, X.; Zhao, X.; Wei, F.; and Wen, J.-R. 2024.
\newblock Language-specific neurons: The key to multilingual capabilities in large language models.
\newblock \emph{arXiv preprint arXiv:2402.16438}.

\bibitem[{Templeton et~al.(2024)Templeton, Conerly, Marcus, Lindsey, Bricken, Chen, Pearce, Citro, Ameisen, Jones, Cunningham, Turner, McDougall, MacDiarmid, Freeman, Sumers, Rees, Batson, Jermyn, Carter, Olah, and Henighan}]{templeton2024scaling}
Templeton, A.; Conerly, T.; Marcus, J.; Lindsey, J.; Bricken, T.; Chen, B.; Pearce, A.; Citro, C.; Ameisen, E.; Jones, A.; Cunningham, H.; Turner, N.~L.; McDougall, C.; MacDiarmid, M.; Freeman, C.~D.; Sumers, T.~R.; Rees, E.; Batson, J.; Jermyn, A.; Carter, S.; Olah, C.; and Henighan, T. 2024.
\newblock Scaling Monosemanticity: Extracting Interpretable Features from Claude 3 Sonnet.
\newblock \emph{Transformer Circuits Thread}.

\bibitem[{Voita, Ferrando, and Nalmpantis(2023)}]{voita2023neurons}
Voita, E.; Ferrando, J.; and Nalmpantis, C. 2023.
\newblock Neurons in large language models: Dead, n-gram, positional.
\newblock \emph{arXiv preprint arXiv:2309.04827}.

\bibitem[{Wang et~al.(2022{\natexlab{a}})Wang, Variengien, Conmy, Shlegeris, and Steinhardt}]{wang2022interpretability}
Wang, K.; Variengien, A.; Conmy, A.; Shlegeris, B.; and Steinhardt, J. 2022{\natexlab{a}}.
\newblock Interpretability in the wild: a circuit for indirect object identification in gpt-2 small.
\newblock \emph{arXiv preprint arXiv:2211.00593}.

\bibitem[{Wang et~al.(2024)Wang, Zhu, Liu, Zheng, Chen, and Li}]{wang2024knowledge}
Wang, S.; Zhu, Y.; Liu, H.; Zheng, Z.; Chen, C.; and Li, J. 2024.
\newblock Knowledge editing for large language models: A survey.
\newblock \emph{ACM Computing Surveys}, 57(3): 1--37.

\bibitem[{Wang et~al.(2022{\natexlab{b}})Wang, Wen, Zhang, Hou, Liu, and Li}]{wang2022finding}
Wang, X.; Wen, K.; Zhang, Z.; Hou, L.; Liu, Z.; and Li, J. 2022{\natexlab{b}}.
\newblock Finding skill neurons in pre-trained transformer-based language models.
\newblock \emph{arXiv preprint arXiv:2211.07349}.

\bibitem[{Ward~Jr(1963)}]{ward1963hierarchical}
Ward~Jr, J.~H. 1963.
\newblock Hierarchical grouping to optimize an objective function.
\newblock \emph{Journal of the American statistical association}, 58(301): 236--244.

\bibitem[{Xiao et~al.(2024)Xiao, Zhang, Song, Jiang, Yao, Han, Wang, Wang, Huang, Lin et~al.}]{xiao2024configurable}
Xiao, C.; Zhang, Z.; Song, C.; Jiang, D.; Yao, F.; Han, X.; Wang, X.; Wang, S.; Huang, Y.; Lin, G.; et~al. 2024.
\newblock Configurable foundation models: Building llms from a modular perspective.
\newblock \emph{arXiv preprint arXiv:2409.02877}.

\bibitem[{Xin et~al.(2025)Xin, Sun, Wang, and Xiong}]{xin2025llmcdsr}
Xin, H.; Sun, Y.; Wang, C.; and Xiong, H. 2025.
\newblock Llmcdsr: Enhancing cross-domain sequential recommendation with large language models.
\newblock \emph{ACM Transactions on Information Systems}.

\bibitem[{Xu, Jain, and Kankanhalli(2024)}]{xu2024hallucination}
Xu, Z.; Jain, S.; and Kankanhalli, M. 2024.
\newblock Hallucination is inevitable: An innate limitation of large language models.
\newblock \emph{arXiv preprint arXiv:2401.11817}.

\bibitem[{Yao et~al.(2023)Yao, Wang, Tian, Cheng, Li, Deng, Chen, and Zhang}]{yao2023editing}
Yao, Y.; Wang, P.; Tian, B.; Cheng, S.; Li, Z.; Deng, S.; Chen, H.; and Zhang, N. 2023.
\newblock Editing large language models: Problems, methods, and opportunities.
\newblock \emph{arXiv preprint arXiv:2305.13172}.

\bibitem[{Zhang and Nanda(2023)}]{zhang2023towards}
Zhang, F.; and Nanda, N. 2023.
\newblock Towards best practices of activation patching in language models: Metrics and methods.
\newblock \emph{arXiv preprint arXiv:2309.16042}.

\bibitem[{Zhang et~al.(2023)Zhang, Zeng, Lin, Xiao, Wang, Han, Liu, Xie, Sun, and Zhou}]{zhang2023emergent}
Zhang, Z.; Zeng, Z.; Lin, Y.; Xiao, C.; Wang, X.; Han, X.; Liu, Z.; Xie, R.; Sun, M.; and Zhou, J. 2023.
\newblock Emergent modularity in pre-trained transformers.
\newblock \emph{arXiv preprint arXiv:2305.18390}.

\bibitem[{Zhao et~al.(2024{\natexlab{a}})Zhao, Chen, Yang, Liu, Deng, Cai, Wang, Yin, and Du}]{zhao2024explainability}
Zhao, H.; Chen, H.; Yang, F.; Liu, N.; Deng, H.; Cai, H.; Wang, S.; Yin, D.; and Du, M. 2024{\natexlab{a}}.
\newblock Explainability for large language models: A survey.
\newblock \emph{ACM Transactions on Intelligent Systems and Technology}, 15(2): 1--38.

\bibitem[{Zhao et~al.(2023{\natexlab{a}})Zhao, Zhang, Ma, Zhang, Gui, Gao, and Huang}]{zhao2023unveiling}
Zhao, J.; Zhang, Z.; Ma, Y.; Zhang, Q.; Gui, T.; Gao, L.; and Huang, X. 2023{\natexlab{a}}.
\newblock Unveiling a core linguistic region in large language models.
\newblock \emph{arXiv preprint arXiv:2310.14928}.

\bibitem[{Zhao et~al.(2023{\natexlab{b}})Zhao, Zhou, Li, Tang, Wang, Hou, Min, Zhang, Zhang et~al.}]{zhao2023survey}
Zhao, W.~X.; Zhou, K.; Li, J.; Tang, T.; Wang, X.; Hou, Y.; Min, Y.; Zhang, B.; Zhang, J.; et~al. 2023{\natexlab{b}}.
\newblock A survey of large language models.
\newblock \emph{arXiv preprint arXiv:2303.18223}, 1(2).

\bibitem[{Zhao et~al.(2024{\natexlab{b}})Zhao, Zhang, Chen, Kawaguchi, and Bing}]{zhao2024large}
Zhao, Y.; Zhang, W.; Chen, G.; Kawaguchi, K.; and Bing, L. 2024{\natexlab{b}}.
\newblock How do large language models handle multilingualism?
\newblock \emph{arXiv preprint arXiv:2402.18815}.

\end{thebibliography}

\clearpage
\appendix
%%% 放到实验设置里去说
% Following \cite{}, we analyze the neurons in the feedforward layers (FFNs) of the LLM, which are crucial for processing and transforming input data. Previous works have shown that these layers are responsible for a wide range of functions~\cite{}.

\section{Experimental Details}
\label{sec_supp:Experimental_Details}

% Our codes are available at 
% \url{https://anonymous.4open.science/r/llm-decoupled-modules}.

\subsection{Dataset}

In this section, we describe the dataset and the preprocessing procedure. The data were sourced from the Infinity-Instruct dataset~\cite{li2025infinity}. Our data selection and categorization followed the methodology of \citet{xiao2024configurable}. We focused on samples that could be unambiguously assigned to one of seven distinct functionalities:

\begin{itemize}
    \item \textbf{Coding:} Various programming languages (e.g., Python, Java, C++), object-oriented programming, code documentation.
    \item \textbf{Math:} Basic calculations to complex problem-solving, mathematical modeling, and proofs.
    \item \textbf{Linguistic:} Syntactic analysis, understanding, and generation.
    \item \textbf{Knowledge:} Diverse subjects like science, literature, geography, and law.
    \item \textbf{Translation:} Multilingual translation, particularly Chinese and English.
    \item \textbf{Ethical:} Ethical reasoning, judgment, and analysis of moral standards.
    \item \textbf{Writing:} Creative writing, scriptwriting, technical writing, narrative generation.
\end{itemize}

We retained only samples that could be exclusively assigned to one of these seven categories. From each category, we randomly selected 1,200 samples, yielding a total of 8,400 samples for our analysis. Of these, 7,000 samples  were randomly chosen as the training set for discovering function modules and training the linear classifier (see Section~\ref{sec:experiments:performance_comparison}). The remaining 1,400 samples were set aside as a hold-out test set to evaluate generalization performance.

\subsection{Definition  of Activation}
% 写一下激活值的定义

In this section, we follow \citet{xiao2024configurable} to define the activation.

Previous works indicate that feedforward layers in the Transformer can be regarded as key-value memory networks~\citep{geva2020transformer} and provide world knowledge for sequence understanding. Therefore, we mainly focus on the feedforward layers for analysis. The feedforward layers (FFNs) employ two-layer projections or gated projections for each token in the sequences. The calculation can be written as $\text{FFN}(\mathbf{x}) = \text{FFN}^O(\text{FFN}^I(\mathbf{x})) = \mathbf{W^O}(\text{FFN}^I(x)) + \mathbf{b^O}$. Here, $\mathbf{W^O} \in \mathbb{R}^{d \times d_{ff}}$ and $\mathbf{b^O} \in \mathbb{R}^d$ are the weight matrix and bias vector for the output linear layer $\text{FFN}^O(\cdot)$. As for $\text{FFN}^I(\cdot)$, there are two variants:
\begin{align*}
    \text{Vallina FFN:} \quad &\text{FFN}^I(\mathbf{x}) = \sigma \left(\mathbf{W^I}\mathbf{x} + \mathbf{b^I}\right), \\
    \text{Gated FFN:} \quad &\text{FFN}^I(\mathbf{x}) = \sigma\left(\mathbf{W_G}\mathbf{x} + \mathbf{b_G}\right) \odot \left(\mathbf{W^I}\mathbf{x} + \mathbf{b^I}\right).
\end{align*}
Here, $\mathbf{W_G}, \mathbf{W^I} \in \mathbb{R}^{d_{\text{ff}}\times d}$ and $\mathbf{b^I}, \mathbf{b_G} \in \mathbb{R}^{d_{\text{ff}}}$ are the weight matrices and bias vectors for the input linear layer $\text{FFN}^I(\cdot)$ and gate linear layer $\text{FFN}_G(\cdot)$. Following previous works~\citep{zhang2023emergent,wang2022finding}, we can split an FFN layer as $d_{\text{ff}}$ neurons, each consisting of a row in the input and gate layer as well as a column in the output layer. The outputs of FFN layers can be rewritten as the sum of all neuron outputs: $\text{FFN}(\mathbf{x}) = \sum_i^{d_{\text{ff}}} \text{FFN}^I(\mathbf{x})_i \mathbf{W}^O_{:,i} + \mathbf{b}^O_i$. We define the intermediate output, $\text{FFN}^I(\mathbf{x})_i$, as the activation value of $i$-th neuron. Intuitively, if the magnitudes of activation values are small, then the corresponding neuron will have a limited impact on the final outputs and vice versa. Therefore, the activation values are widely used as indicators of neuronal functionality.

We denote the LLM to be analyzed as $\mathcal{M}$, which has $L$ Transformer layers and $L\times d_{ff}$ neurons. Given a neuron $n$, as the FFN layers are computed in a token-wise manner, we can collect the activation values of the neuron $n$ on the collection $\mathcal{C}$. Then, the collected activation of neuron $u_i$ for sample $s_j$ at token $t$ is denoted as $a_{i,j,t}$. The average activation magnitude of neuron $u_i$ on sample $s_j$ is then defined as the mean of the absolute normalized activations across all tokens:
$A_{i,j} = \frac{1}{T_j} \sum_{t=1}^{T_j} |a_{i,j,t}|$, where $T_j$ is the number of tokens in sample $s_j$.
Besides, to ensure comparable activation magnitudes across neurons, each neuron's average activation magnitude is normalized by \textit{z-score} over all samples.

To ensure that activation magnitudes are comparable across different neurons, we then normalize these scores using \textit{z-score}. For each neuron $u_i$, we compute the mean $\mu_{A_i}$ and standard deviation $\sigma_{A_i}$ of its average activation magnitudes over all samples. The final normalized activation for neuron $u_i$ on sample $s_j$ is: 
\begin{equation}
A'_{i,j}= \frac {A_{i,j} - \sigma_{A_i}}{\sigma_{A_i}},
\end{equation}
which serves as the indicator for the neuron's functionality on a given sample.

\begin{figure*}[t!]
    \vspace{-0.3cm}
    \centering
    \includegraphics[width=\textwidth]{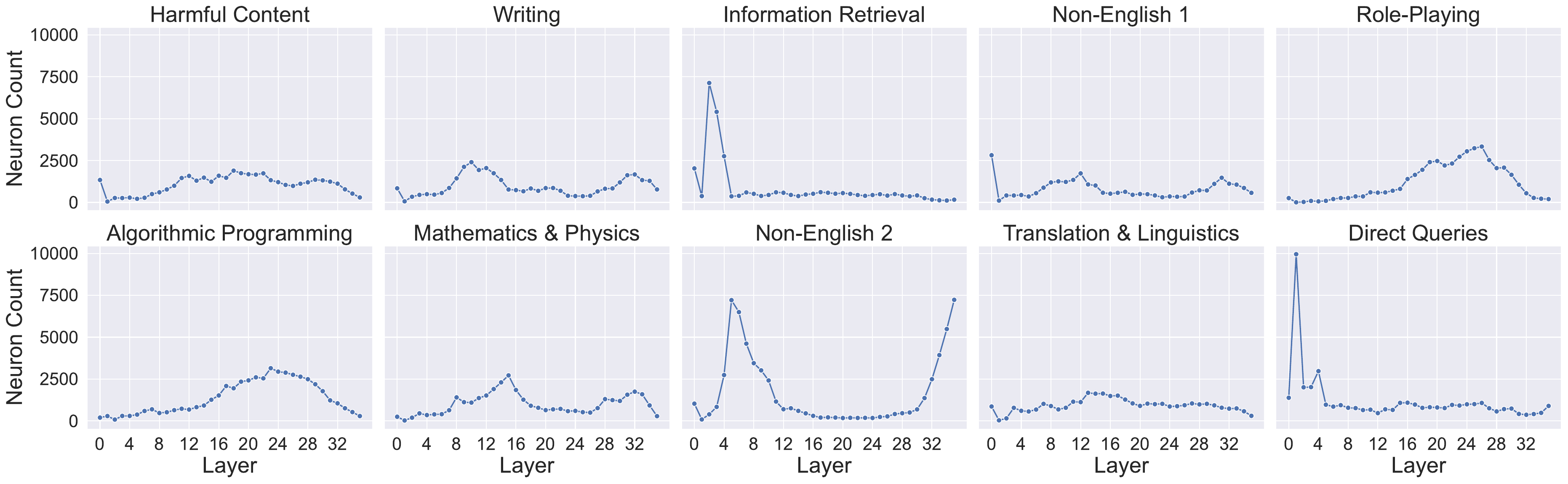}
    \caption{Layer-wise distribution of function modules in Qwen2.5-3B-Instruct ($K=10$). Each subfigure shows the neuron counts in each layer for the corresponding function module.}
    \label{fig:layer_dist}
    \vspace{-0.1cm}
\end{figure*}

\section{Explanation of IterD}
\label{sec_supp:Explanation_of_IterD}

In this section, we present a line-by-line explanation of the IterD algorithm for identifying function modules, as detailed in Algo.~\ref{alg:algorithm}.

\begin{description}
    \item[Line 1] IterD begins by creating an initial partition $F_0$ of all samples and neurons into $K$ groups. This partition can be initialized randomly or by using a clustering method as a heuristic starting point.

    \item[Line 2] An iteration counter $t$ is initialized to $0$.

    \item[Lines 3--17] These lines constitute the main iterative loop. The algorithm repeatedly alternates between updating neuron assignments and sample assignments to monotonically increase the objective function $\mathcal{L}(F)$.

    \begin{description}

    \item[Line 4] At the beginning of each iteration, the current partition $F_t$ is unpacked into its sample and neuron groups $\{(S_k^t, U_k^t)\}_{k=1}^K$.

    \item[Lines 5--9] This block performs the first step: \textbf{optimizing neuron assignments}.
    
        \begin{description}
            \item[Line 5] A temporary copy of the current neuron partition, $\mathcal{P}'_U = \{U'_1,\dots,U'_K\}$, is created and initialized as $\{U_k^t\}_{k=1}^K$.

            \item[Line 6] Iterating over every neuron $u \in U$.

            \item[Line 7] For each neuron $u$, it evaluates the objective obtained by hypothetically moving $u$ to each of the $K$ modules. $\mathcal{L}(F_{u \to k})$ denotes the value of the objective if neuron $u$ were reassigned to module $k$. The best module index is
            \begin{equation}
            k_u^* \leftarrow \arg\max_{k \in \{1,\dots,K\}} \mathcal{L}(F_{u \to k}).
            \end{equation}

            \item[Line 8] Neuron $u$ is then moved to module $k_u^*$ in the temporary partition $\mathcal{P}'_U$. 

        \end{description}
    
        \item[Lines 10--14] This block performs the second step: \textbf{optimizing sample assignments}.

        \begin{description}
            \item[Line 10] Using the updated neuron groups $\{U'_k\}_{k=1}^K$ from the previous step, a temporary sample partition $\mathcal{P}'_S = \{S'_1,\dots,S'_K\}$ is created and initialized as the current sample groups, i.e., $\mathcal{P}'_S = \{S_k^t\}_{k=1}^K$.

            \item[Line 11] Iterating over every sample $s \in S$.

            \item[Line 12] For each sample $s$, it greedily determines the best module assignment
            \begin{equation}
            k_s^* \leftarrow \arg\max_{k \in \{1,\dots,K\}} \mathcal{L}(F_{s \to k}).
            \end{equation}

            \item[Line 13] The sample $s$ is then reassigned to module $k_s^*$ in the temporary sample partition $\mathcal{P}'_S$.

        \end{description}

    \item[Line 15] After the previous two steps, the updated neuron and sample partitions are combined to form the new partition $F_{t+1} \leftarrow \{(S'_k, U'_k)\}_{k=1}^K$.
    
    \item[Line 16] The iteration counter is incremented, $t \leftarrow t + 1$, so that $t$ now indexes the most recently constructed partition.

    \item[Line 17] This line implements the \emph{convergence check}. The loop terminates when the current partition $F_t$ is identical to the partition from the previous iteration $F_{t-1}$. %In other words, once a full pass over all neurons and all samples produces no changes in assignments, the algorithm has converged to a stable solution.

    \end{description}
    
    \item[Line 18] After convergence, the algorithm returns the final, stable partition $F_{t}$ as the discovered function modules.
    
\end{description}

\section{Function Determination via LLM}
\label{sec_supp:Function_Determination_via_LLM}

In this section, we provide the details of how we obtain the specific functions of the extracted function modules via LLMs. (recalling Sec.~\ref{subsec:Empirical_Insights_of_Function_Modules}). 

% add some motivations here.

First, for each sample $s_j$ and each function module $(S_k, U_k)$, we compute the mean activation value of $s$ on $U_k$, \ie, $\mathcal{A}(s_j,\  U_k)=\frac{1}{|U_k|}  \sum_{u_i \in U_k} A_{i,j}$, recalling the definition of the matrix $A$ in Sec.~\ref{subsec:Optimization_Objectives}.
% For each function module $U_k$, we compute its activation value $A(s, U_k)$ for every sample $s$ in our dataset.
Then, for a given function module $(S_k, U_k)$, We identify the set of $H$ samples $S'_k = \{s_{k,0}, s_{k,1}, \dots, s_{k,H-1}\}$ ($|S'_k| = H$) that yield the highest $H$ activation values for $(S_k, U_k)$.
Finally, these samples for each module are then compiled into a single prompt and sent to an advanced LLM for qualitative analysis.

We send these samples to the LLM for analysis with the following prompt structure:

\newtcolorbox{prompt}[2][]
  {colback = blue!3!white,
   colframe = blue!70!black,
   fonttitle = \bfseries,
   colbacktitle = blue!80!black,
   coltitle = white,
   enhanced,
   attach boxed title to top center={yshift=-2mm},
   title=#2,#1}

\begin{prompt}[]{Prompt Structure}
I analyzed the activation patterns of the neurons in LLMs and obtained some sample categories. Analyze the commonalities of each category. Here are these samples:

--- category 0: ---

$s_{0,0}$

$s_{0,1}$

$s_{0,2}$

...

$s_{0,H-1}$

--- category 1: ---

$s_{1,0}$

$s_{1,1}$

$s_{1,2}$

...

$s_{1,H-1}$

...

--- category $K-1$: ---

$s_{K-1,0}$

$s_{K-1,1}$

$s_{K-1,2}$

...

$s_{K-1,H-1}$
\end{prompt}

The LLM's response provides a summary of the commonality for each category, which is then used to assign a function label to the corresponding module $(S_k, U_k)$.

A response example for the five categories is as follows: 

\newtcolorbox{response}[2][]
  {colback = green!3!white,
   colframe = green!70!black,
   fonttitle = \bfseries,
   colbacktitle = green!80!black,
   coltitle = white,
   enhanced,
   attach boxed title to top center={yshift=-2mm},
   title=#2,#1}

\begin{response}[]{Response}
Okay, I can help you analyze the commonalities of these sample categories. Here's an analysis of each one:

---

\textbf{Category 0: Task \& Persona Specification for Language Models}

The samples in \textbf{Category 0} overwhelmingly focus on defining a specific persona or task for a large language model. Many prompts explicitly state "I want you to act as..." or "You are the text completion model..." followed by detailed instructions on the desired output format, tone, and constraints.

Essentially, this category represents prompts where the user is trying to precisely control the model's behavior and output for a defined task.

---

\textbf{Category 1: Short, Direct Queries}

\textbf{Category 1} is characterized by short, direct inputs. These samples are generally concise and appear to be initiating a conversation or requesting information in a specific context.

This category seems to group simple, often one-line, requests or commands.

---

\textbf{Category 2: Non-English Language Content \& Task}

The defining characteristic of \textbf{Category 2} is that the majority of the samples are in languages other than English. These prompts involve various tasks, including questions, requests for translation, or text completion, all within a non-English context.

While the tasks vary (e.g., answering questions, translation, content generation), the common thread is the use of a non-English language for the primary interaction.

---

\textbf{Category 3: Creative Writing \& Narrative Generation}

\textbf{Category 3} is strongly focused on creative writing, particularly narrative and screenplay generation. The prompts request detailed, imaginative content with specific stylistic and thematic elements.

This category is clearly about generating fictional and creative text with a strong emphasis on storytelling and character.

---

\textbf{Category 4: Algorithmic \& Competitive Programming Problems}

\textbf{Category 4} consists almost entirely of problem descriptions from competitive programming platforms or algorithmic challenges. These samples typically include an input format, output format, examples, and a detailed problem statement requiring a logical or mathematical solution, often implemented via code.

This category is a collection of well-defined algorithmic tasks, likely used for training or evaluating models on problem-solving and code generation capabilities.
\end{response}

\section{Layer Sensitivity of Functions within LLMs}
\label{sec_supp:Layer_Sensitivity_of_Functions_within_LLMs}

In this section, we further analyzed the layer sensitivity of functions across all layers of Qwen2.5-3B-Instruct, as illustrated in Fig.~\ref{fig:layer_dist}. Different functions exhibit distinct neuron distribution patterns, indicating that specific functionalities are localized to different depths within the model. These patterns align with two primary trends also identified in \citet{lindsey2025biology}.

First, cognitive simple functions are concentrated in the early-to-mid layers. For example, \textit{Information Retrieval} and \textit{Direct Queries} show a sharp peak in neuron count at layers 1 and 2. It suggests the model's foundational capacity for identifying and processing information is handled within the initial layers.

Second, functions demanding complex reasoning rely heavily on the mid-to-late layers. \textit{Algorithmic Programming} peaks at layer 23, while \textit{Mathematics \& Physics} peaks at layers 15 and 32. Similarly, \textit{Role-Playing}, which requires sustained contextual understanding and persona consistency, peaks at layer 26. This pattern implies that the model's more advanced cognitive abilities, such as logical deduction and abstract thought, emerge from the deeper layers of its architecture.

Second, functions demanding complex reasoning rely heavily on the mid-to-late layers. \textit{Algorithmic Programming} peaks at layer 23, while \textit{Mathematics \& Physics} peaks at layers 15 and 32. Similarly, \textit{Role-Playing} which requires sustained contextual understanding and persona consistency, peaks at layer 26. This pattern implies that the model's more advanced cognitive abilities, such as logical deduction and abstract thought, emerge from the deeper layers of its architecture.

Furthermore, some functions exhibit more distributed profiles. \textit{Writing} shows broad activation across the middle layers, and \textit{Translation \& Linguistics} demonstrates a relatively flat distribution, suggesting these functions may involve processing across multiple stages rather than being highly localized. Notably, the \textit{Non-English 2} is concentrated in the early and late layers, with minimal presence in the middle layers. It suggests that for non-English queries, the model may follow a non-English to English-centric and back to non-English processing pipeline, a phenomenon also mentioned by ~\citet{zhao2024large}.

\end{document}